\documentclass[10pt,twocolumn,letterpaper]{article}

\usepackage{cvpr}
\usepackage{times}
\usepackage{epsfig}
\usepackage{graphicx}
\usepackage{amsmath}
\usepackage{amssymb}
\usepackage{url}
\usepackage{subfig}
\usepackage{wrapfig}
\usepackage[linesnumbered,algoruled,boxed,lined]{algorithm2e}

\cvprfinalcopy 
\def\cvprPaperID{1327}

\ifcvprfinal\fi
\begin{document}

\title{Deep Neural Networks In Fully Connected CRF For Image Labeling With Social Network Metadata}

\author{Chengjiang Long ~~~~~~ Roddy Collins ~~~~~~ Eran Swears ~~~~~~ Anthony Hoogs\\
Kitware Inc.\\
28 Corporate Drive, Clifton Park, NY 12065\\
{\tt\small \{chengjiang.long, roddy.collins, eran.swears, anthony.hoogs\}@kitware.com}
}

\maketitle

\begin{abstract}

We propose a novel method for predicting image labels by fusing image content descriptors with the social media context of each image. An image uploaded to a social media site such as Flickr often has meaningful, associated information, such as comments and other images the user has uploaded, that is complementary to pixel content and helpful in predicting labels. Prediction challenges such as ImageNet~\cite{imagenet_cvpr09} and MSCOCO~\cite{LinMBHPRDZ:ECCV14} use only pixels, while other methods make predictions purely from social media context \cite{McAuleyECCV12}. Our method is based on a novel fully connected Conditional Random Field (CRF) framework, where each node is an image, and consists of two deep Convolutional Neural Networks (CNN) and one Recurrent Neural Network (RNN) that model both textual and visual node/image information. The edge weights of the CRF graph represent textual similarity and link-based metadata such as user sets and image groups. We model the CRF as an RNN for both learning and inference, and incorporate the weighted ranking loss and cross entropy loss into the CRF parameter optimization to handle the training data imbalance issue. Our proposed approach is evaluated on the MIR-9K dataset and experimentally outperforms current state-of-the-art approaches.
\end{abstract}

\section{Introduction}

Multimedia data such as images and videos are being produced and shared at an unprecedented and accelerating pace in recent years. For example, on YouTube, video data is currently being uploaded at the rate of approximately 30 million hours a year. This drives a strong need to develop automatic tools to help users understand, organize, and retrieve images and videos from vast collections. While recent advances have been impressive, real-world multimedia, especially those shared on the image-sharing platform Flickr, can still be challenging to index and retrieve using only visual information, due to complex content, partial occlusion, and diverse styles and quality.



\begin{figure}
\begin{center}
\includegraphics[height=0.70\linewidth, width=0.99\linewidth, angle=0]{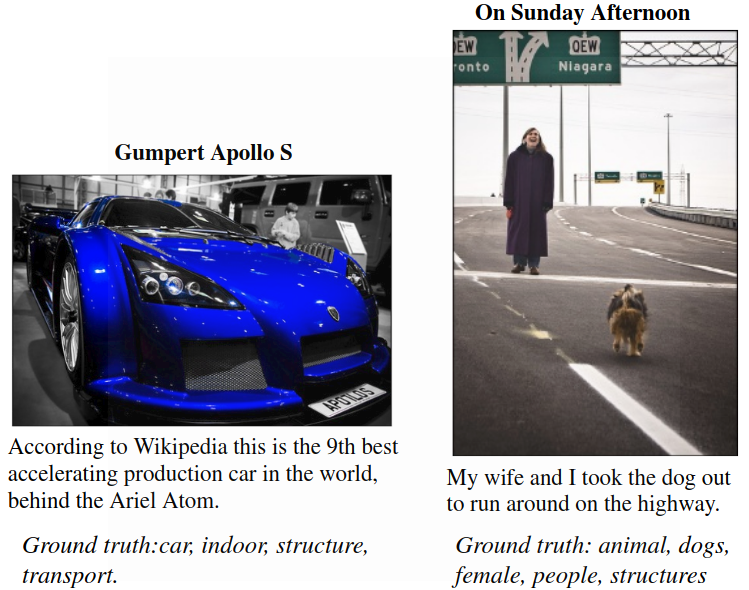}
\end{center}
\vspace{-0.40cm}
\caption{Two sample images with the title in bold, image description and the corresponding ground-truth labels in italic from the MIR-9K dataset. The goal of this paper is to make full use of such text information as well as the link-based metadata like user sets and image groups to boost the quality of image labeling.}
\label{fig:nfdd_problem}
\end{figure}


Images in social media do not exist in isolation. As illustrated in Figure~\ref{fig:nfdd_problem}, a rich social multimedia database contains images, text information such as image title, description and comments, as well as user information ({\em e.g.}, username, location, network of contacts), user image gallery, uploader-defined groups, and links between shared content. Most image recognition and label prediction methods depend entirely or primarily on pixel content, and do not make full use of commonly-available multimedia information to aid in automatic image labeling. We hypothesize that using social media context jointly with pixel information should improve the state-of-the-art in image labeling. Furthermore, we seek to understand the relative contribution of pixels, text and other information in predicting image labels.


We define our problem as an automatic image labeling based on inferring content labels $Y$, conditioned on an image $I$, and other related metadata information $M$. Our proposed solution is illustrated in Figure~\ref{fig:scrfcnn_pipeline}, which introduces a novel deep fully connected Conditional Random Field framework (we call ``DCRF") that uses deep neural networks to compute the joint probability $P(Y|I,M)$. CRFs have been commonly used in image segmentation problems where the model has one hidden node per pixel or grid-cell, and a vector of hidden nodes for a single image. Instead, we abstract up one layer and define one hidden node per image, instead of per pixel, over the entire dataset of images to form an image relationship graph. This results in having the vector of hidden nodes over the whole dataset with one node per image.

For pixel content descriptors, we exploitation popular image classification CNNs to extract a visual feature vector for each image or CRF node. To incorporate image title, comments, captions and other text, instead of using high-frequency words as tags~\cite{McAuleyECCV12, Johnson:ICCV15}, we treat the text information as an unorganized and incoherent sentence and then fine-tune a popular network for sentence classification~\cite{Kim14f} as a text-level neural network to extract text features. In addition to textual similarity based on the text feature, we use associative metadata such as user sets and image groups to determine the edge weights in the fully connected CRF graph.

Our fully connected CRF establishes pairwise potentials on all pairs of images over the entire dataset. It combines the strengths of both CNN and CRF based graphical models in a unified framework. Inspired by Zheng {\em et al.}~\cite{Zheng:ICCV2015}, we formulate a mean-field approximation inference~\cite{krahenbuhl2011efficient} for the CRF and model it as a Recurrent Neural Network (RNN). Hence our DCRF is an end-to-end CNN-RNN framework, incorporating the advantages of both convolutional and recurrent neural networks, while enabling standard back-propagation during training for network parameter learning.


In the most closely related work, McAuley {\em et al.}~\cite{McAuleyECCV12} has proposed a CRF framework using social-network metadata to solve the image labeling problem. Compared with McAuley's approach, we have two advantages. First, our image-level CNN makes full use of the existing popular CNN models to extract powerful visual features from images, which integrates of the advantages of CNN feature extraction for nodes in the CRF.
Second, rather than exploring the relational model based on high-frequency co-occurring words as tags, we exploit our text-level CNN and associative metadata to construct the fully connected CRF graph. The experiment section shows that our method results in significant performance improvement. 


Our main contributions are summarized as follows:
\begin{itemize}
\item We propose a novel deep fully connected CRF framework DCRF that uses deep neural networks for image labeling with social network metadata. Deep CCN image features are fused with text features and network linkage information in an end-to-end deep learning formulation. 
\item Instead of using high-frequency words as tags, we propose to use a text-level CNN to exploit textual information. The fully connected CRF graph is built based on the features extracted from the text-level convolutional neural network, as well as the link-based metadata like user sets and image groups.
\item For both learning and inference, we model a mean field approximation inference~\cite{krahenbuhl2011efficient} for the fully connected CRF as an RNN, to introduce the CNN-RNN formulation. We also incorporate the weighted ranking loss to handle the imbalance label distribution existing in the training data.
\item We evaluate the proposed DCRF on the MIR-9K dataset and achieve significantly improved performance compared to previous state-of-the-art approaches.
\end{itemize}

\vspace{-0.05cm}
\section{Related work}
The related work can be divided into two categories: {\em social media for labeling} and {\em CRF with deep neural networks}.

{\bf Social media context for labeling}. A set of tags associated with each image is commonly used in multimodal classification settings. Guillamumin {\em et al.}~\cite{guillaumin2010} explored the relationship between tags and manual annotations to recover annotations using a combination of tags and image content. Lindstaedt {\em et al.}~\cite{Lindstaedt2008} and Sigurbjornsson {\em et al.}~\cite{sigurbjoernsson2008} studied the problem of recommending tags that were obtained from similar images and similar users. Sawant {\em et al.}~\cite{SawantDLW10} and Stone {\em et al.}~\cite{Stone2008} investigates friendship information between users for tag recommendation in social networks. EXIF and GPS are two commonly used sources of metadata that come directly from the camera~\cite{patil2017survey, LiCH09, Kalogerakis2009, JoshiLYLG11}. Such metadata can be used to help determine who captured the photo and where, and also provide informative signals for image labeling tasks. Our method differs from all these and also \cite{McAuleyECCV12} in that we use a much larger range of social media information, including free-form text as well as links, with deep learning based pixel descriptors incorporated into our novel deep learning fully connected CRF framework.

{\bf CRF with deep neural networks}. In recent years, there are several works about CRF with a convolutional neural network which incorporate CRF to model structures in both output and hidden feature layers in CNN. Chu Chao {\em et al.}~\cite{ChuOLW16} propose a CRF-CNN framework which can simultaneously model structural information in both output and hidden feature layers in a probabilistic way, and apply it to human pose estimation. Shuai Zheng {\em et al.}~\cite{crfasrnn_iccv2015} introduce a new form of convolutional neural network that combines the strengths of CNN and CRF-based probabilistic graphical modeling. Zheng {\em et al.}~\cite{Zheng:ICCV2015} models conditional random fields for image segmentation task as recurrent neural networks, with the node features extracted from a convolutional neural network. Chandra {\em et al.}~\cite{Chandra_2017_ICCV} propose a structured prediction model that endows the Deep Gaussian Conditional Random Field with a densely connected graph structure. In these works, CNNs are integrated into CRF models and perform as feature extractors. Similarly, the two CNNs (image-level CNN and text-level CNN) in our proposed DCRF framework also work as powerful feature extractors in image labeling using social metadata. However, in addition to this, our text-level CNN is also used to help build the fully connected CRF graph, and both the learning and inference is under the united CNN-RNN framework, which distinguishes our proposed DCRF from the existing approaches.

\section{Proposed approach}
\begin{figure*}
\begin{center}
\includegraphics[height=0.30\linewidth, width=0.78\linewidth, angle=0]{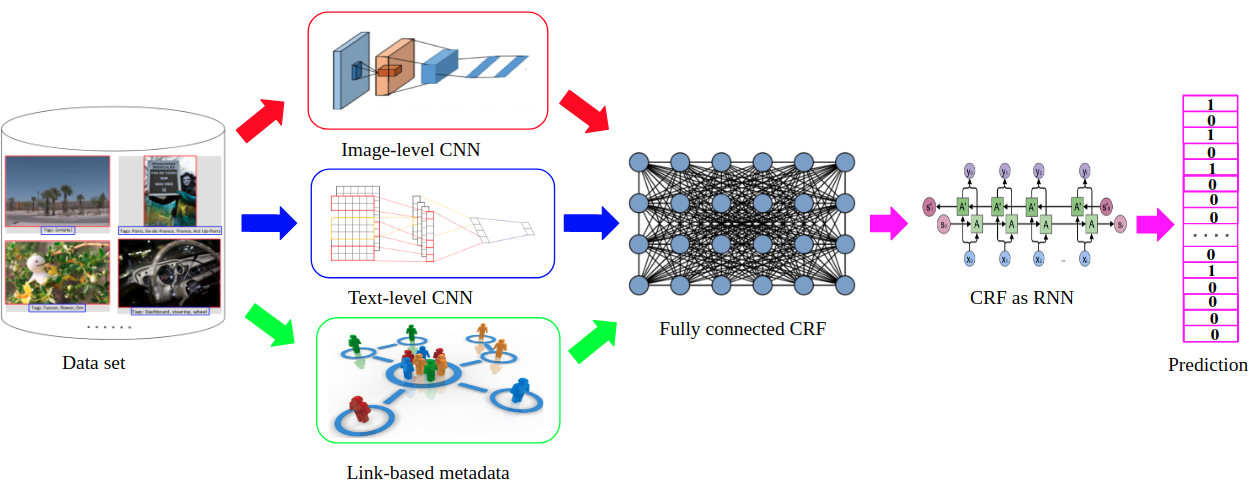}
\end{center}
\caption{The pipeline of the proposed Deep fully connected CRF framework with deep neural networks for image labeling using social network metadata. The node features are extracted from an image-level convolutional neural network (CNN), and the edges are built based on the textual similarity based on the feature extracted from a text-level CNN, as well as the similarity determined by link-based metadata like user sets and image groups. For both learning and inference, we resort to a mean field approximation inference~\cite{krahenbuhl2011efficient} for the CRF and model it as an RNN. We learn the parameter via a stochastic gradient descent RMSProp under the united CNN-RNN framework. Based on the learned CNN-RNN framework, we can predict final 24-dimensional binary label vector directly at the testing stage.}
\label{fig:scrfcnn_pipeline}
\vspace{-0.30cm}
\end{figure*}

In this section, we will describe in detail the proposed Deep fully connected CRF framework (as illustrated in Figure~\ref{fig:scrfcnn_pipeline}) with deep neural networks.

\subsection{CRF framework}
Our probability framework is based on a fully connected conditional random field (CRF). This captures both unary dependencies between image labels, $Y=\{y_1, y_2, \ldots, y_N\}$ (with binary value indicating if the image has this class label, $y_n = 1$ , or not, $y_n = 0$), and the input features ({\em e.g.}, image features and metadata), as well as the pairwise dependencies between pairs of labels and the input features to produce the conditional probability $P(Y|I,M) = P(Y|{\bf x}, M)$, where ${\bf x}$ are the raw image features derived from the image set $I$. 
The labels are treated as binary hidden nodes in the CRF and the image features ${\bf x}$, and metadata $M$, are used in the observation nodes. Therefore, the conditional probability of the fully connected CRF can be defined as:
\begin{equation}
\begin{split}
P(Y|I,M) &= P(Y|{\bf x}, M) \\
&= \frac{1}{Z}\exp(\sum\limits_{i=1}^NA(y_i, {\bf x}_i) \\
&\quad+ \sum\limits_{i=1}^N\sum\limits_{\forall j \neq i} B(y_i, y_j, M)),\label{eqn:objectivefunction}
\end{split}
\end{equation}
where $Z$ is the normalization constant that depends on ${\bf x}$ and $M$, while $A$ is the unary function based on the image information ${\bf x}$, and $B$ is the pairwise potential function based on the metadata $M$. The unary potentials are single image potentials, while the pairwise potentials are between pairs of images. 
For simplicity, a separate binary CRF model can be learned for each category. 



\subsection{Unary function with image-level convolutional neural network}\label{sec:img-cnn}
The goal of the image-level convolutional neural network (CNN) is to extract feature vectors that are compact, representative, and can capture the most related visual information for the decoder. The rapid development of deep convolutional neural networks have had great success in large-scale image recognition task~\cite{He_2016_CVPR}, object detection~\cite{NIPS2015_5638} and visual captioning~\cite{VenugopalanXDRMS14}. High-level features can be extracted from upper or intermediate layers of a deep CNN network. Therefore, a set of well-tested CNN networks can be used in our framework. 

We use VGG-19~\cite{Simonyan14c} and ResNet-152 network~\cite{He_2016_CVPR} for our framework. In this paper, for each category, we modify the original network by changing the number of outputs in the last layer from 1000 to 2 and fine-tune to conduct the binary classification. ${\bf x}_i$ in Equation~\ref{eqn:objectivefunction} is the feature vector extracted from the second last fully connected layer ({\em i.e.}, the 18-th layer in VGG-19 and the 151-th layer in ResNet-152) for $i$-th instance. Then we can define the unary potential function as
\begin{equation}
A(y_i, {\bf x}_i) = {\bf w}_A^{y_i} {\bf x}_i + {\bf b}_A^{y_i},
\end{equation}
where $y_i$ is either 1 or 0, and ${\bf w}_A^{y_i}$, ${\bf b}_A^{y_i}$ are the parameters we need to learn.

\subsection{Pairwise potential with text-level convolutional neural network and other meta information}\label{sec:tagcnn}
Unlike previous work that tries to make full use of text information in the metadata by exploring the co-occurrence of high frequently used words as tags, we treat all the texts including title, description and comments information associated with an image as an unorganized incoherent sentence or a bag of words. Then we can train a text-level convolutional neural network to extract the feature vectors. In principle, any sentence convolutional neural networks can be used in our framework. To make it simple, we use Kim's sentence network~\cite{Kim14f}, which is composed of one convolutional layer, one pooling layer, one dropout layer, one fully-connected layer and output with softmax activation function. 

In this paper, we extract the 128-dimensional dropout layer to measure the similarity between any two images at the text-level. We define text similarity as
\begin{equation}
S_{text}(i, j) = \exp(-\frac{|{\bf x}_i^{text}-{\bf x}_j^{text}|^2}{2\theta_{text}}),\label{eqn:simtex}
\end{equation}
where the degree of nearness and similarity is controlled by the $\theta_{text}$ parameter.

Besides the text information, we also use the link-based metadata such as user sets and image groups~\cite{Johnson:ICCV15}. A user set associates with a collection of images uploaded or collected by the same user. Image groups are community-curated, and are usually images belonging to the same concept, scene or event are uploaded and shared by the social network users. Both user sets and image groups have vocabularies, {\em i.e.}, $T_{set}$ and $T_{group}$, and each image ${\bf x}_i$ has two subsets, $t_i^{set}$ and $t_i^{group}$. We calculate the distance between any two nodes/images using the Jaccard similarity between their user sets and image groups as:
\begin{equation}
d(i, j, M_{set}) = 1 - |t_i^{set} \cap t_j^{set}|/|t_i^{set} \cup t_j^{set}|, \nonumber
\end{equation}
\begin{equation}
d(i, j, M_{group}) = 1 - |t_i^{group} \cap t_j^{group}|/|t_i^{group} \cup t_j^{group}|  \nonumber
\end{equation} 
and get the corresponding similarities 
\begin{equation}
S_{set}(i, j) = \exp(-\frac{d(i, j, M_{set})^2}{2\theta_{set}}),
\end{equation}
\begin{equation}
S_{group}(i, j) = \exp(-\frac{d(i, j, M_{group})^2}{2\theta_{group}}),
\end{equation}
where both $\theta_{set}$ and $\theta_{group}$ are the parameter to control the degree of similarity.

Intuitively, two nodes/images which are more similiar are more likely to share the same labels and should be able to affect each other more than others. Therefore, similarity is close related to the pairwise potential and we can define the pairwise potential as 
\begin{equation}
B(y_i, y_j, M)) =  \mu(y_i, y_j)\sum\limits_{k\in \{text, set, group\}}{\bf w}_B^k S_k(i, j),
\end{equation}
where $\mu(\cdot,\cdot)$ is label compatibility function, and ${\bf w}_B^{text}$, ${\bf w}_B^{set}$, ${\bf w}_B^{group}$ are the $2\times2$ parameters to be learned from the training data.


\subsection{Learning and inference under neural network}\label{sec:learnandinference}
Both image-level CNN and text-level CNN are trained separately. With the pre-trained networks, we are able to extract the node features and textual features to build the fully connected CRF, in which we also take user sets and image groups into account. Unfortunately, the parameters $\theta_{text}$, $\theta_{set}$ and $\theta_{group}$ cannot be calculated efficiently since their gradients involve a sum of non-Gaussian kernels, which are not amenable to the same acceleration techniques. Therefore, we resort to take grid search on a holdout validation set for determining all these three parameters, and learn the parameter ${\bf w} = [{\bf w}_A, {\bf w}_B]$, ${\bf b}_A$ and $\mu(\cdot,\cdot)$ only.

\begin{algorithm}
 \caption{The outline of our proposed DCRF algorithm}\label{algo:outline}
 \KwIn{$I$ and $M$}
 \KwOut{$Q$}
 ${\bf x}  \leftarrow \text{CNN}_{image}(I)$ \\
 ${\bf x}^{text}  \leftarrow \text{CNN}_{text}(M)$ \\
 ${\bf t}^{set}, {\bf t}^{group}  \leftarrow M$ \\
 $U \leftarrow {\bf w}_A {\bf x} + {\bf b}_A $ \\
 $Q_i(y) \leftarrow \frac{1}{Z_i}\exp{\{U_i(y)\}}$ \\ 
 \While{not converged}{ 
  ${\tilde Q}_i^{(k)}(y) \leftarrow \sum\limits_{\forall j \neq i}S_k(i, j)Q_j(y)$ for all $k$ \\ 
  ${\check{Q}}_i(y) \leftarrow \sum\limits_{k}{\bf w}_B^k {\tilde Q}_i^{(k)}(y)$ \\ 
  ${\hat Q}_i(y) \leftarrow \sum\limits_{y^{\prime}}\mu(y, y^{\prime}){\check{Q}}_i(y)$ \\ 
  ${\breve{Q}}_i(y) \leftarrow U_i(y) - {\hat Q}_i(y)$ \\ 
  $Q_i(y) \leftarrow \frac{1}{Z_i}\exp{\{{\breve{Q}}_i(y)\}}$ \\ 
 }
\end{algorithm}

We resort to a mean field approximation inference~\cite{krahenbuhl2011efficient} which computes a distribution $Q({\bf X})$ that minimizes the KL-divergence $D(Q||P)$ among all the approximated distributions $Q$ that can be experessed as a product of independent marginals, $Q({\bf X}) = \prod_i Q_i$ \footnote{For simplicity, we use $Q_i$ to indicate $Q({\bf x}_i)$ and therefore $Q_i(y)$ indicate the probability of ${\bf x}_i$ being labeled as label $y$.} and
\begin{equation}
\begin{split}
Q_i(y) = \frac{1}{Z_i}\exp{\{-{\bf w}_A^y {\bf x}_i - Q_i^{\prime}(y)\}},\label{eqn:mfaprox_function}
\end{split}
\end{equation}
where
\begin{equation}
Q_i^{\prime}(y) = \sum\limits_{y^{\prime}}\mu(y, y^{\prime})\sum\limits_{k\in \{text, set, group\}}{\bf w}_B^k \sum\limits_{\forall j \neq i}S_k(i, j)Q_j(y). \nonumber
\end{equation}

The update equation in Equation~\ref{eqn:mfaprox_function} leads to the inference steps, as seen in Algorithm~\ref{algo:outline}. Inspired by the spirits in Zheng {\em et al.}'s work~\cite{Zheng:ICCV2015}, we can implement the algorithm as a combination framework with both CNN and RNN. To be specific, line 1 and line 2 are associated with image-level CNN and text-level CNN, respectively. Line 3 can be regarded as a preprocessing step. Line 4 can be modeled as a fully connected layer. Line 5 is a softmax layer with unary potential as input. Line 7 can be regarded as linear combination of matrix multiplications, since the parameters $\theta_{text}$, $\theta_{set}$ and $\theta_{group}$ are determined by grid search validation and therefore $S_k(\cdot,\cdot)$ is fixed during running the algorithm~\ref{algo:outline}. Line 8 can be implemented as a convolution with a 1x1 filter with three input channels and one output channel. Line 9 is another convolutional layer in which the number of both input and output channels are both two for the binary classification case. Line 10 is element-wise subtraction from the unary potential $U_i$. Line 11 is another softmax layer. Obviously, the layers associated with line 7 to line 11 construct a recurrent neural network (RNN).

Both learning and inference can be conducted under the united CNN-RNN framework which we implement in PyTorch. For the loss function, besides the L2 regularization on ${\bf w}_A$, we use weighted binnary cross entropy and pairwise ranking loss 
\begin{equation}
\begin{split}
&Loss(Q, Y) = \sum\limits_{i=1}^N -\frac{y_i}{N_{+}}\log{Q_i(y_i)} -\frac{1-y_i}{N_{-}}\log{Q_i(1-y_i)} \\
&\quad\quad+\sum\limits_{i=1}^N \frac{y_i}{N_{+}}(1 - (Q_i(y_i)-Q_i(1-y_i)) \\
&\quad\quad+\sum\limits_{i=1}^N \frac{1-y_i}{N_{-}}(1 - (Q_i(1-y_i)-Q_i(y_i)) + \lambda{\left\lVert{\bf w}_A\right\rVert}_2, 
\end{split}\label{eqn:lossfunc}
\end{equation}
to handle the possible imbalance distribution of positive/negative instances in the training data and ensure a good probability ranking. Note that $N_{+}$ and $N_{-}$ in Equation~\ref{eqn:lossfunc} are the number of postive training instances and the number of the negative training instances, respectively. $\lambda$ is the regularization parameter and we set 0.1 for VGG-19 and 0.001 for ResNet-152. We initialize all parameters using the method of~\cite{KaimingHeZRS:ICCV15} and optimize using stochastic gradient descent RMSProp with a fixed learning rate of 0.1.

\begin{figure}
\begin{center}
\includegraphics[height=0.35\textwidth, width=0.38\textheight]{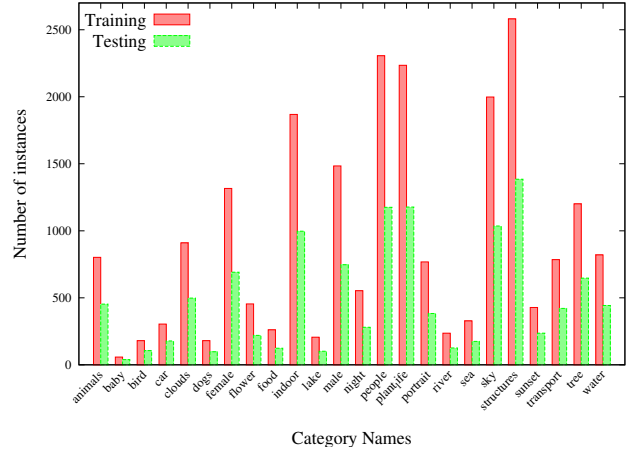}
\end{center}
\vspace{-0.55cm}
\caption{The number of instances per category used for both training and testing among 24 categories on the full MIR-9K dataset.}\label{fig:datadistribution}
\vspace{-0.35cm}
\end{figure}

\vspace{-0.10cm}
\section{Experiment}
\vspace{-0.15cm}
We conduct experiments to verify the effectiveness of the proposed approach on the MIR-9K dataset, a subset of the MIRFLICKR~\cite{HuiskesMIR08} dataset which is available under Creative Commons licenses. 
It worths mentioning here that we do not evaluate on datasets like NSU-WIDE dataset~\cite{Chua09nus-wide:CIVR09} since there are only tag words available on the official website, without the original text information ({\em e.g.}, image title, description and comments), which prevent us from evaluating our text-level CNN in Section~\ref{sec:tagcnn}. The MIR-9K dataset contains 6000 training instances and 3182 testing instances with 24 categories: animals, baby, bird, car, clouds, dogs, female, flower, food, indoor, lake, male, night, people, plant life, portrait, river, sea, sky, structures, sunset, transport, tree, and water. 
It involves a set 3,213 users, a collection of 34,942 words and 17,687 image groups.
The data distribution 
is shown in Figure~\ref{fig:datadistribution}, where there are imbalance issues among different categories. 

For measurement metrics, we report the average precision (AP), recall, precision and accuracy over all 24 categories for the sake of comparison with published algorithms. 
\begin{figure*}
\begin{minipage}{0.245\textwidth}
\begin{center}
\subfloat[AP]{\label{fig:tagcnnvgg_ap}
\includegraphics[height=0.58\textwidth, width=0.20\textheight, angle=0]{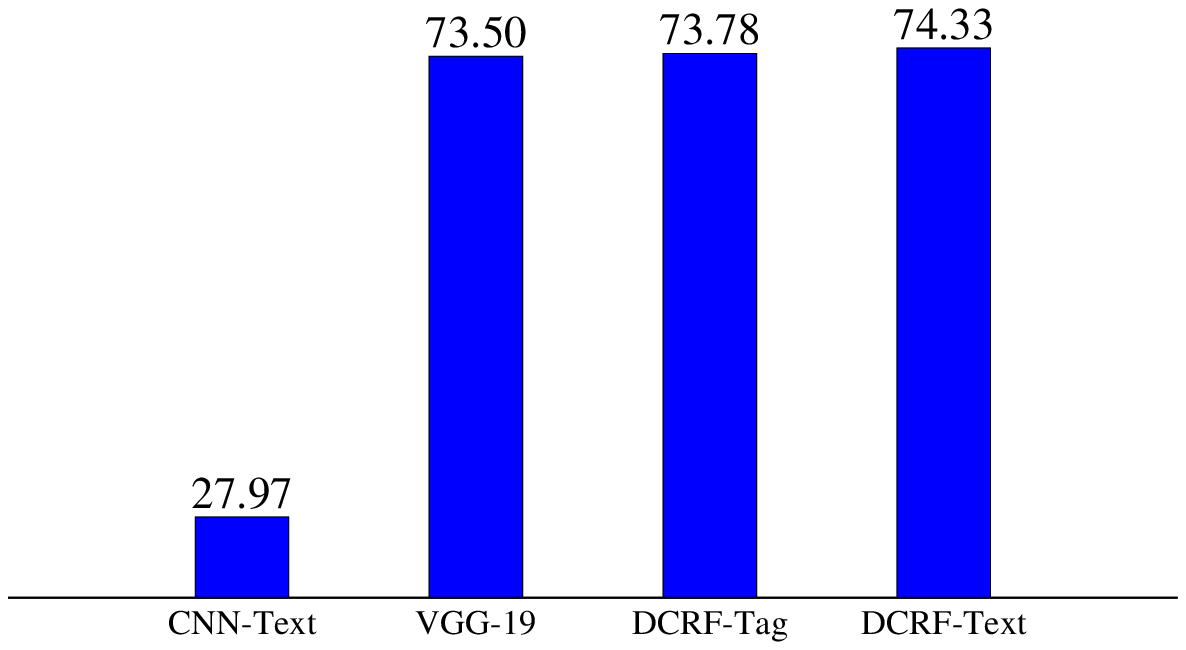}}
\end{center}
\end{minipage}
\begin{minipage}{0.245\textwidth}
\begin{center}
\subfloat[Recall]{\label{fig:tagcnnvgg_rec}
\includegraphics[height=0.58\textwidth, width=0.20\textheight, angle=0]{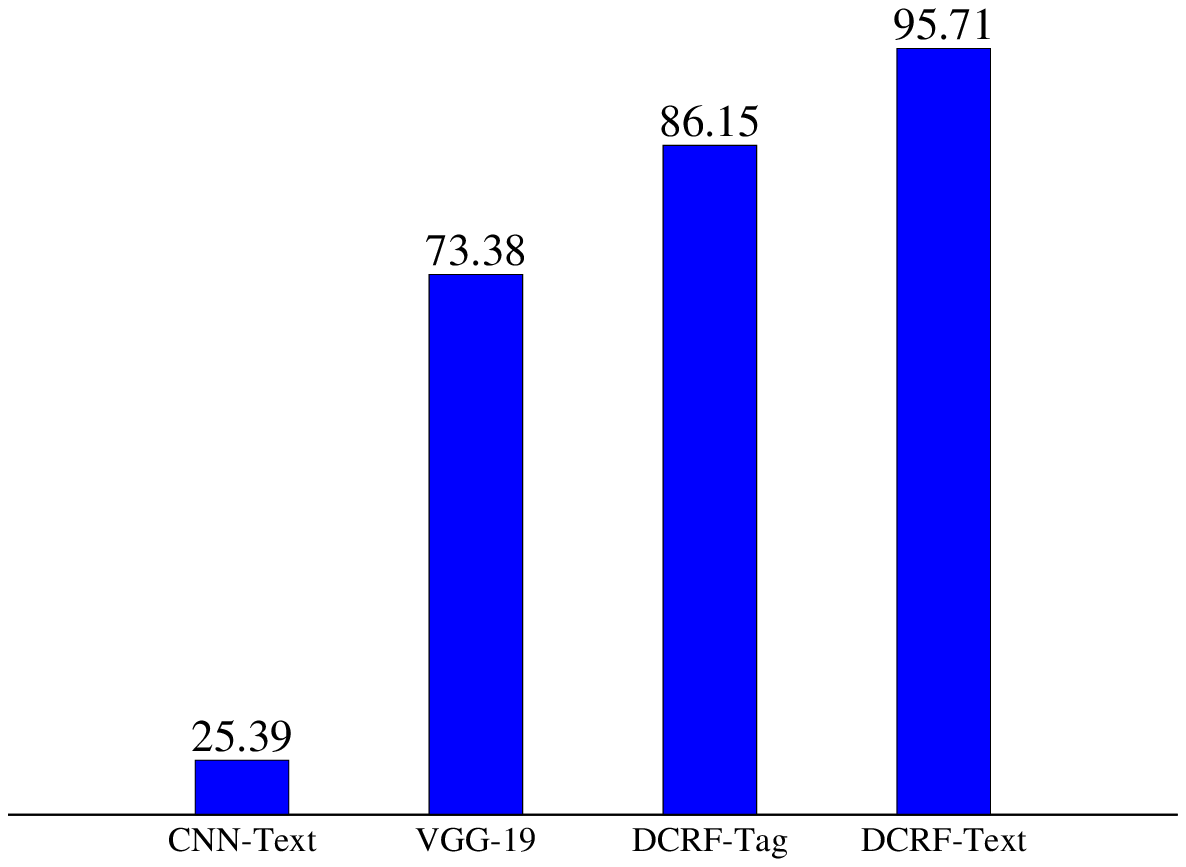}}
\end{center}
\end{minipage}
\begin{minipage}{0.245\textwidth}
\begin{center}
\subfloat[Precision]{\label{fig:tagcnnvgg_pre}
\includegraphics[height=0.58\textwidth, width=0.20\textheight, angle=0]{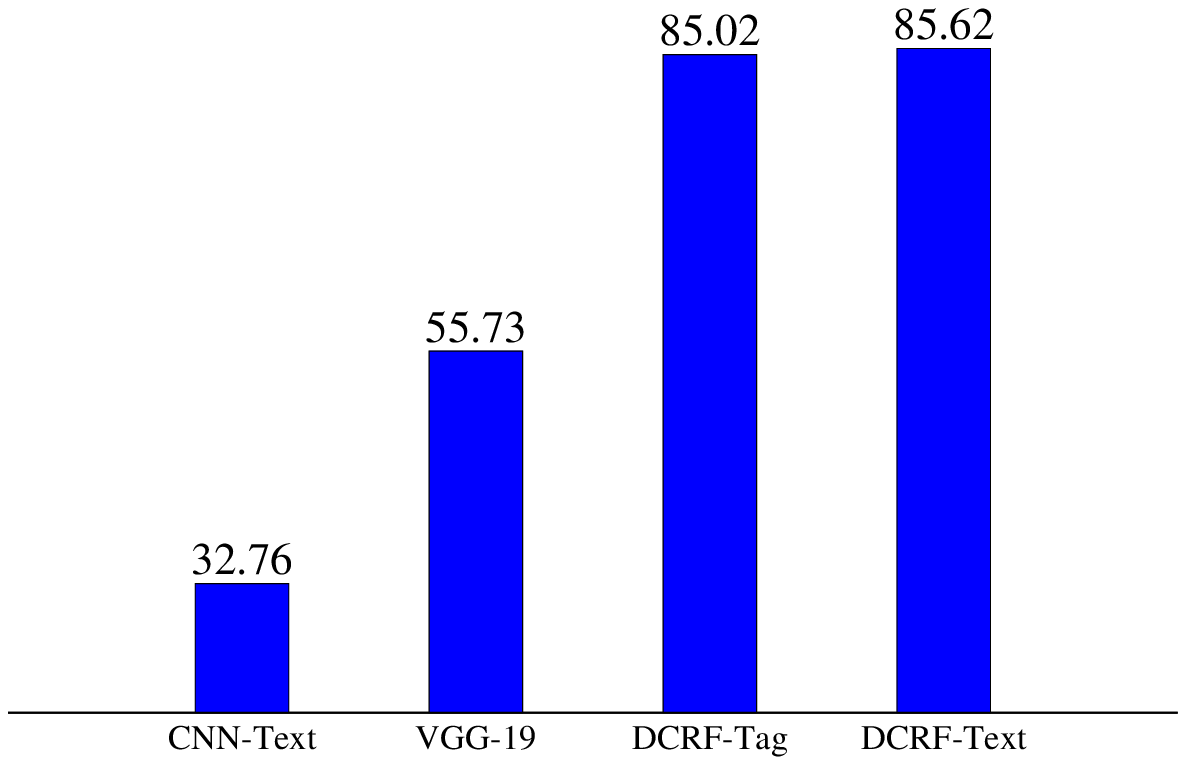}}
\end{center}
\end{minipage}
\begin{minipage}{0.245\textwidth}
\begin{center}
\subfloat[Accuracy]{\label{fig:tagcnnvgg_acc}
\includegraphics[height=0.58\textwidth, width=0.20\textheight, angle=0]{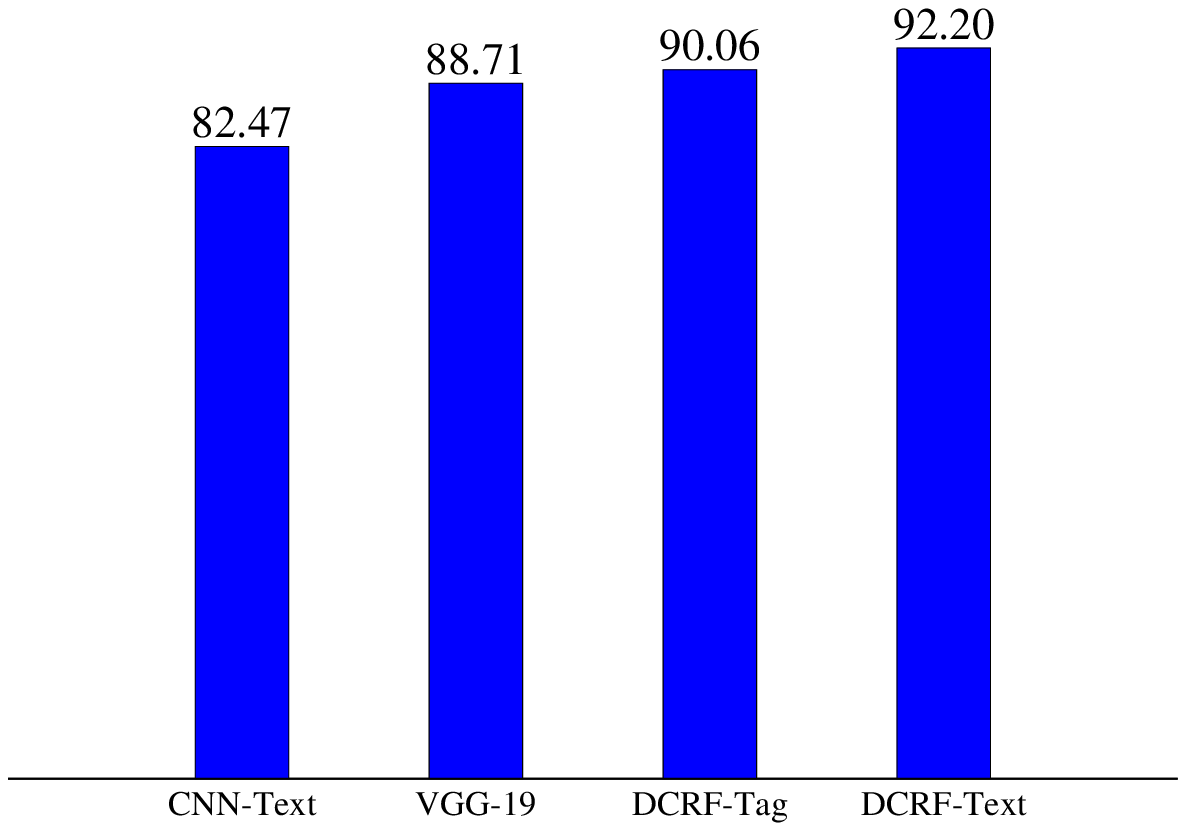}}
\end{center}
\end{minipage}~
\vspace{-0.10in}
\caption{The comparison result with node features extracted from the VGG-19 network on the MIR-9K dataset (unit: \%).}\label{fig:tag_vggcnn}
\vspace{-0.20in}
\end{figure*}
\begin{figure*}
\begin{minipage}{0.245\textwidth}
\begin{center}
\subfloat[AP]{\label{fig:tagcnnres_ap}
\includegraphics[height=0.58\textwidth, width=0.20\textheight, angle=0]{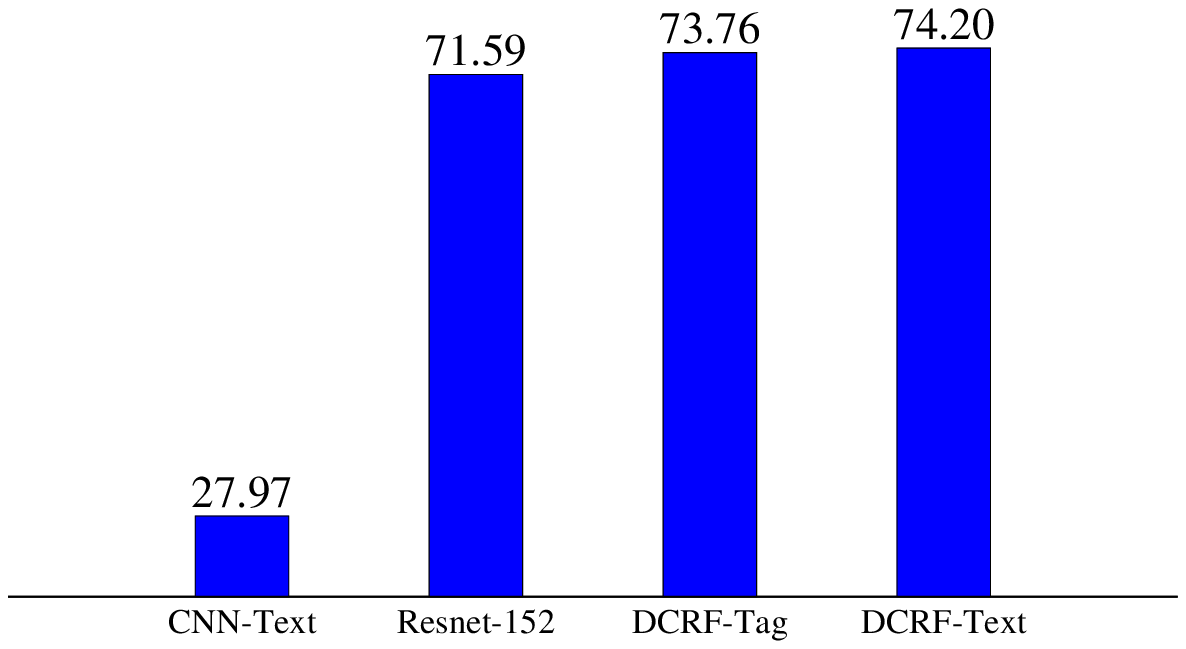}}
\end{center}
\end{minipage}
\begin{minipage}{0.245\textwidth}
\begin{center}
\subfloat[Recall]{\label{fig:tagcnnres_rec}
\includegraphics[height=0.58\textwidth, width=0.20\textheight, angle=0]{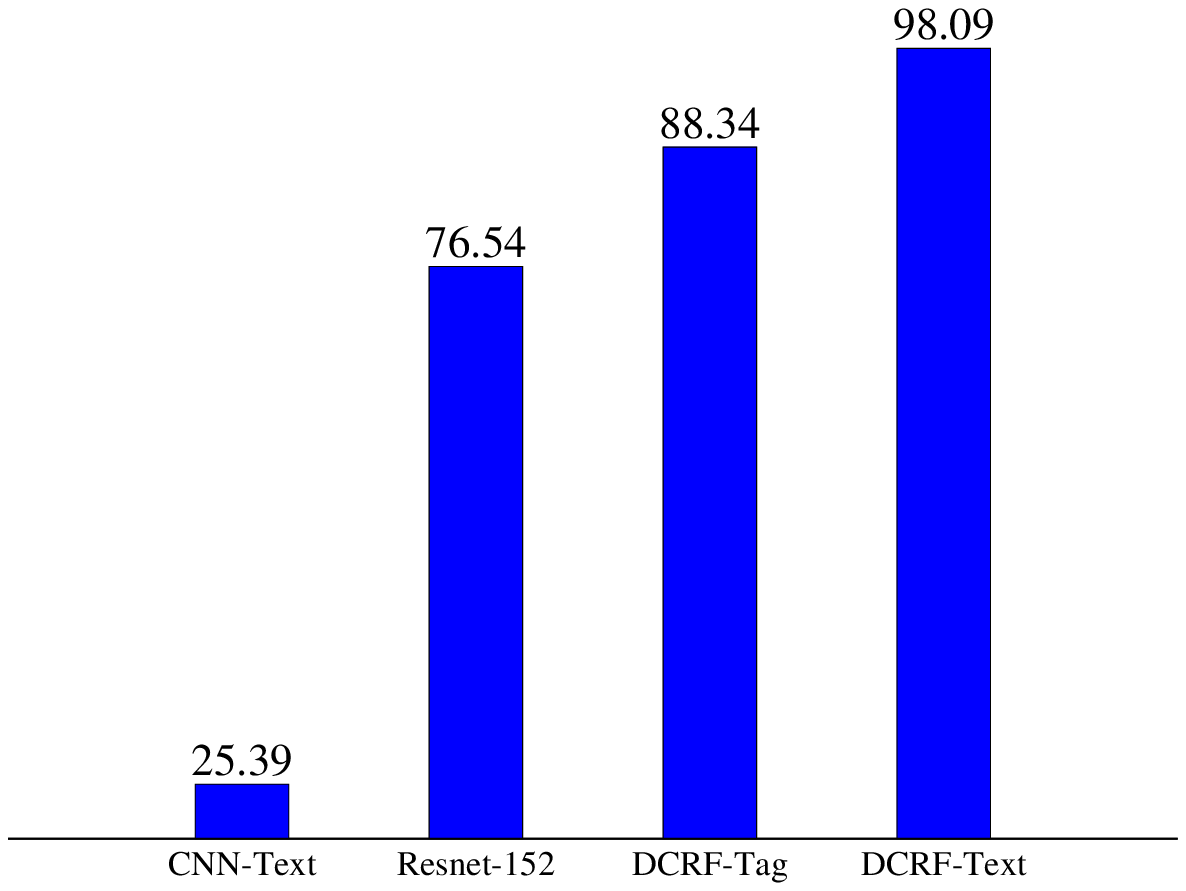}}
\end{center}
\end{minipage}
\begin{minipage}{0.245\textwidth}
\begin{center}
\subfloat[Precision]{\label{fig:tagcnnres_pre}
\includegraphics[height=0.58\textwidth, width=0.20\textheight, angle=0]{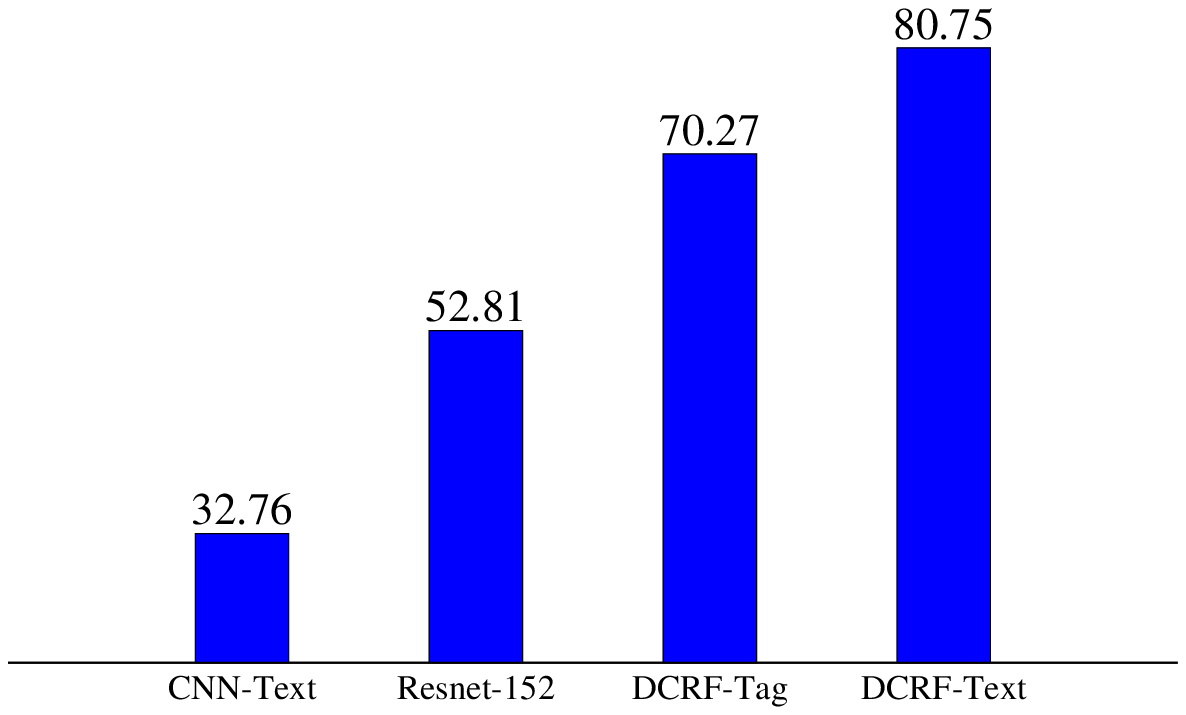}}
\end{center}
\end{minipage}
\begin{minipage}{0.245\textwidth}
\begin{center}
\subfloat[Accuracy]{\label{fig:tagcnnres_acc}
\includegraphics[height=0.58\textwidth, width=0.20\textheight, angle=0]{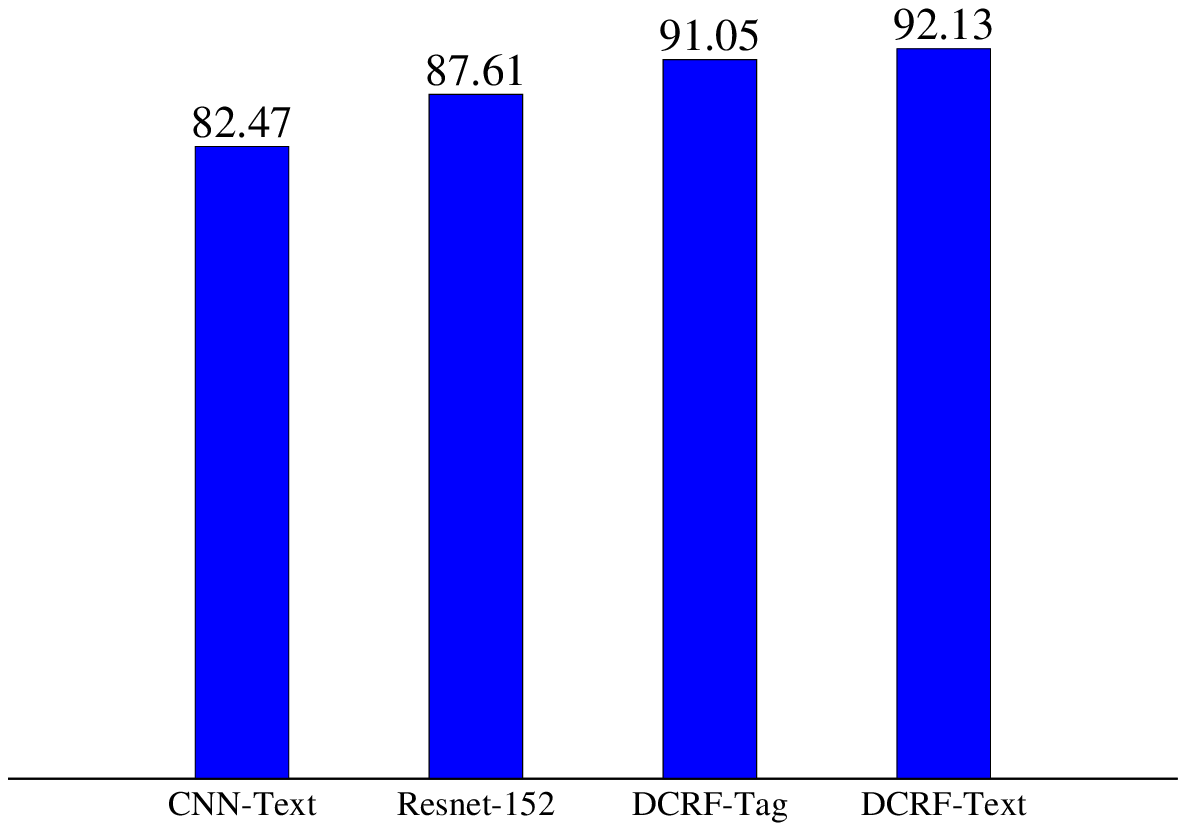}}
\end{center}
\end{minipage}~
\vspace{-0.10in}
\caption{The comparison result with node features extracted from the ResNet-152 network on the MIR-9K dataset (unit: \%).}\label{fig:tag_rescnn}
\vspace{-0.20in}
\end{figure*}
\begin{figure}
\captionsetup[subfigure]{labelformat=empty}
\begin{minipage}{0.230\textwidth}
\begin{center}
\subfloat[animal]{\label{fig:tagclass0}
\includegraphics[height=0.70\textwidth, width=0.18\textheight, angle=0]{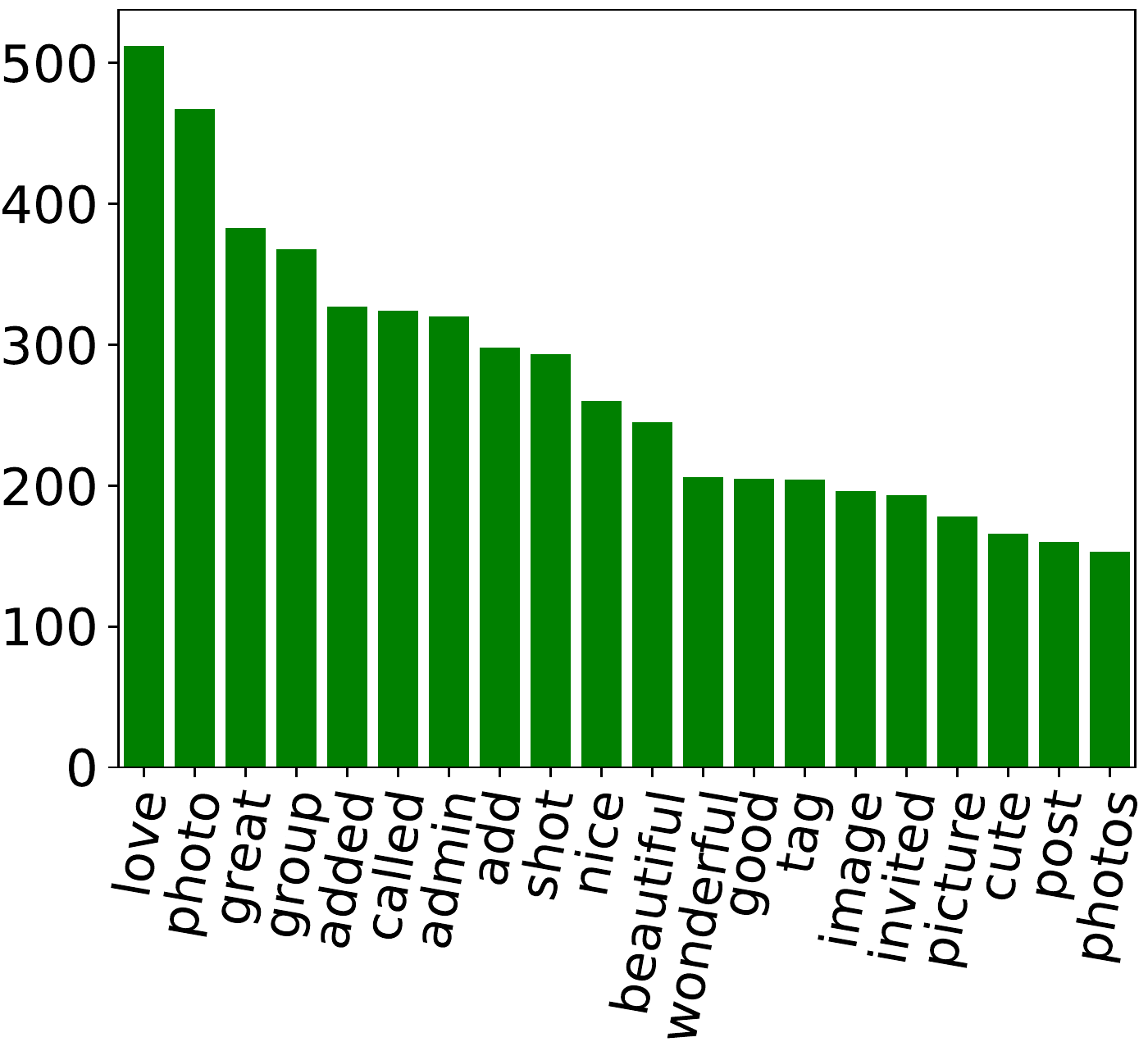}}
\end{center}
\end{minipage}
\begin{minipage}{0.230\textwidth}
\begin{center}
\subfloat[flower]{\label{fig:tagclass7}
\includegraphics[height=0.70\textwidth, width=0.18\textheight, angle=0]{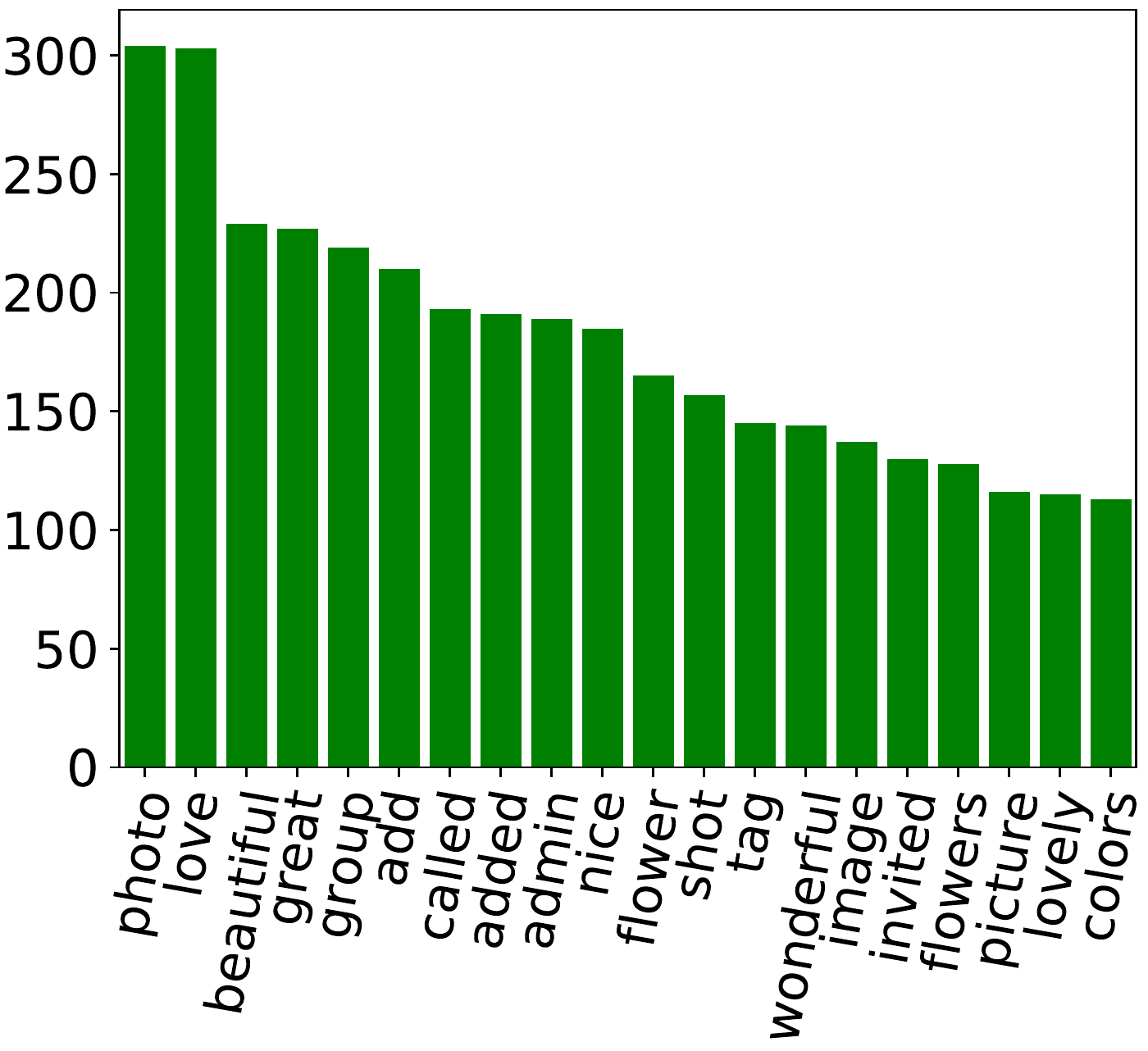}}
\end{center}
\end{minipage}\\
\begin{minipage}{0.230\textwidth}
\begin{center}
\subfloat[portrait]{\label{fig:tagclass15}
\includegraphics[height=0.70\textwidth, width=0.18\textheight, angle=0]{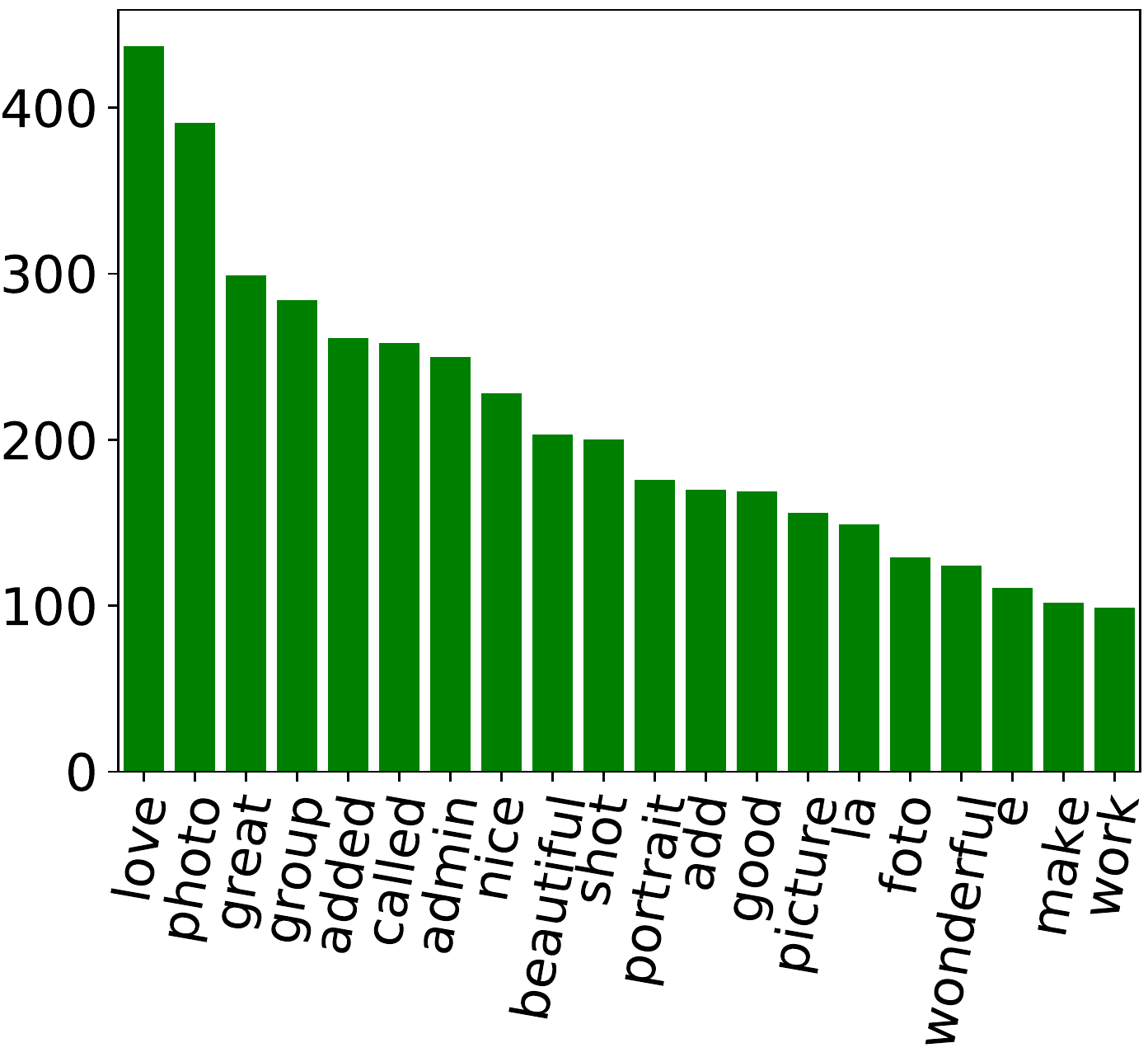}}
\end{center}
\end{minipage}
\begin{minipage}{0.230\textwidth}
\begin{center}
\subfloat[water]{\label{fig:tagclass23}
\includegraphics[height=0.70\textwidth, width=0.18\textheight, angle=0]{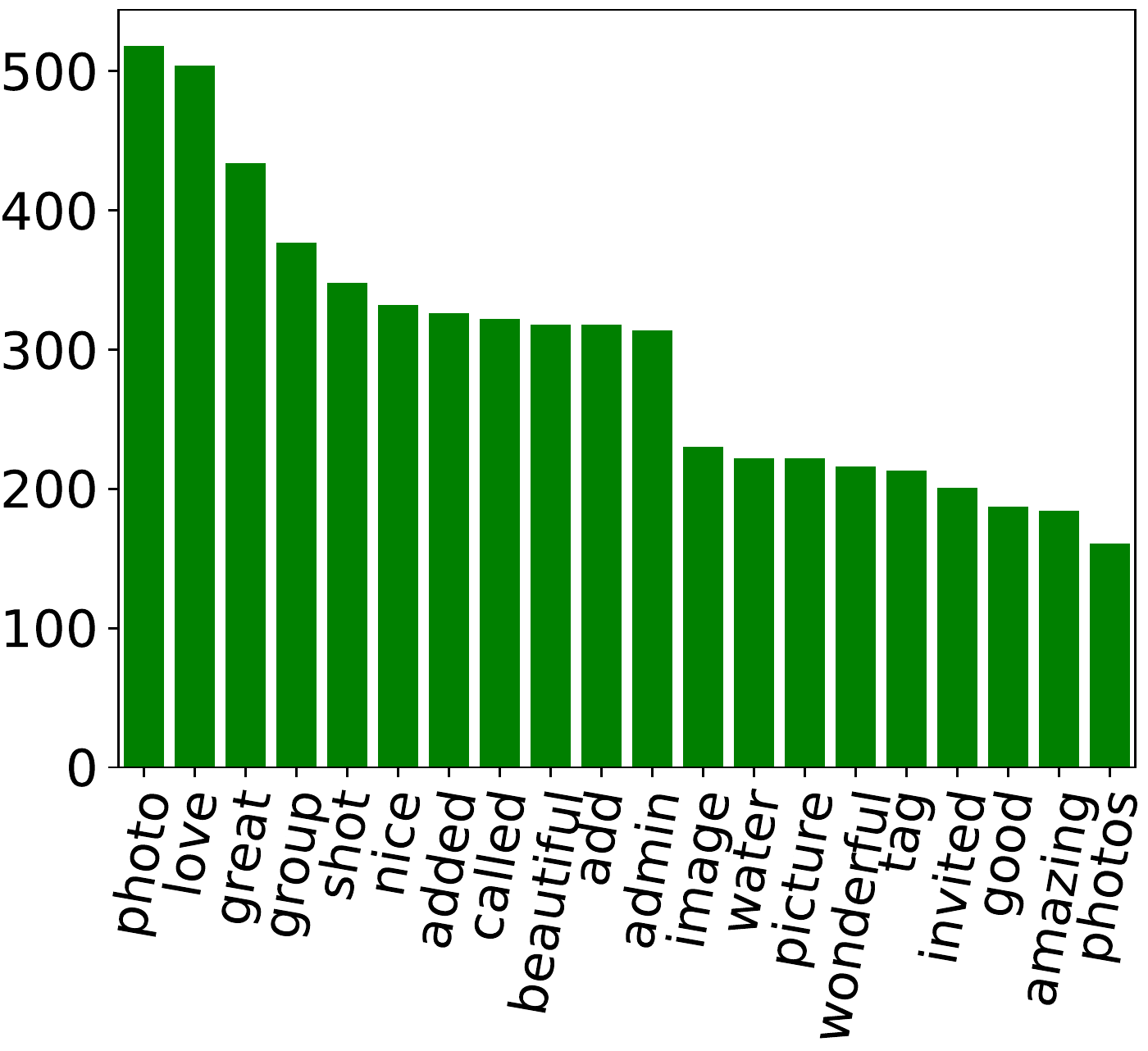}}
\end{center}
\end{minipage}~
\vspace{-0.10in}
\caption{Visualization of top 20 tag words appearing in 4 categories on the positive training instances of the MIR-9K dataset.}\label{fig:toptagwordshistogram}
\vspace{-0.20in}
\end{figure}

\subsection{Effectiveness of the text-level CNN}\label{sec:effect_tag_cnn}
To evaluate the effectiveness of the text-level CNN to our proposed DCRF, we first define the CRF only with the textual similarity defined in Equation~\ref{eqn:simtex}, and denote the method as DCRF-Text. We then replace the textual similarity with the tag Jaccard similarity calcualted in ~\cite{Johnson:ICCV15} (with 5000 high frequent occurring words
as tags) for any two nodes/images to build the CRF, and mark the competing algorithm as DCRF-Tag. Note that both DCRF-Text and DCRF-Tag use the same visual node features obtained from the VGG-19 or ResNet-152 network. We also provide two baselines that use text information and image information independently, {\em i.e.}, CNN-Text, VGG-19 and ResNet-152, respectively.

The performance results are summarized in Figure~\ref{fig:tag_vggcnn} and~\ref{fig:tag_rescnn}. As we can see, (a) both VGG-19 and ResNet-152 perform much better than CNN-Text in AP, recall, precision and accuracy, which indicates image information is more helpful for image labeling when compared with the text information. (b) Also both DCRF-Text and DCRF-Tag work better than CNN-Text, VGG-19 and ResNet-152 in all four metrics, which indicates the text information complementary to image information and useful for improving the labeling accuracy. (c) Regardless of using the VGG-19 or ResNet-152 network to extract the node/image features, DCRF-Text outperforms DCRF-Tag, which clearly demonstrate that the text-level CNN of our DCRF is better able to explore the underlying information in text than just relying on the top frequent words as tags. 

To analyze why our proposed DCRF-Text is able to outperform DCRF-Tag, we visualize the top 20 words in decreasing order of frequency occurring among 6000 training examples. Due to space limitations, we only present 4 categories in Figure~\ref{fig:toptagwordshistogram}. 
 We observe that the top frequently co-occurring words such as ``love", ``photo", ``great", ``group", 
``added", ``nice" {\em et. al} convey little information relative to any of the prediction 24 categories.  Instead, we resort to a text-level CNN to explore the underlying information and use the textual similarity based on the features extracted from the text-level CNN to build the fully connected CRF graph, which explains why our proposed approach DCRF makes slightly better use of text information.

\subsection{Effectiveness of the metadata for image labeling}
In Section~\ref{sec:effect_tag_cnn}, text information has been proved to able to boost the performance for the image labeling accuracy. We also want to see whether the link-based metadata such as user sets and image groups, can produce positive effect on the image labeling. Using the same node features extracted from either the VGG-19 or ResNet-152 network, we first define the CRF graph with a single type of metadata ({\em i.e.}, text, user sets and image groups) and get three versions of DCRF: DCRF-Text, DCRF-Set and DCRF-Group. Then we define the CRF with the combined these three types of metadata together, and denote the combined version as DCRF-TSG.
\begin{figure}[h]
\begin{minipage}{0.230\textwidth}
\begin{center}
\subfloat[AP with VGG-19]{\label{fig:metacnnvgg_ap}
\includegraphics[height=0.58\textwidth, width=0.20\textheight, angle=0]{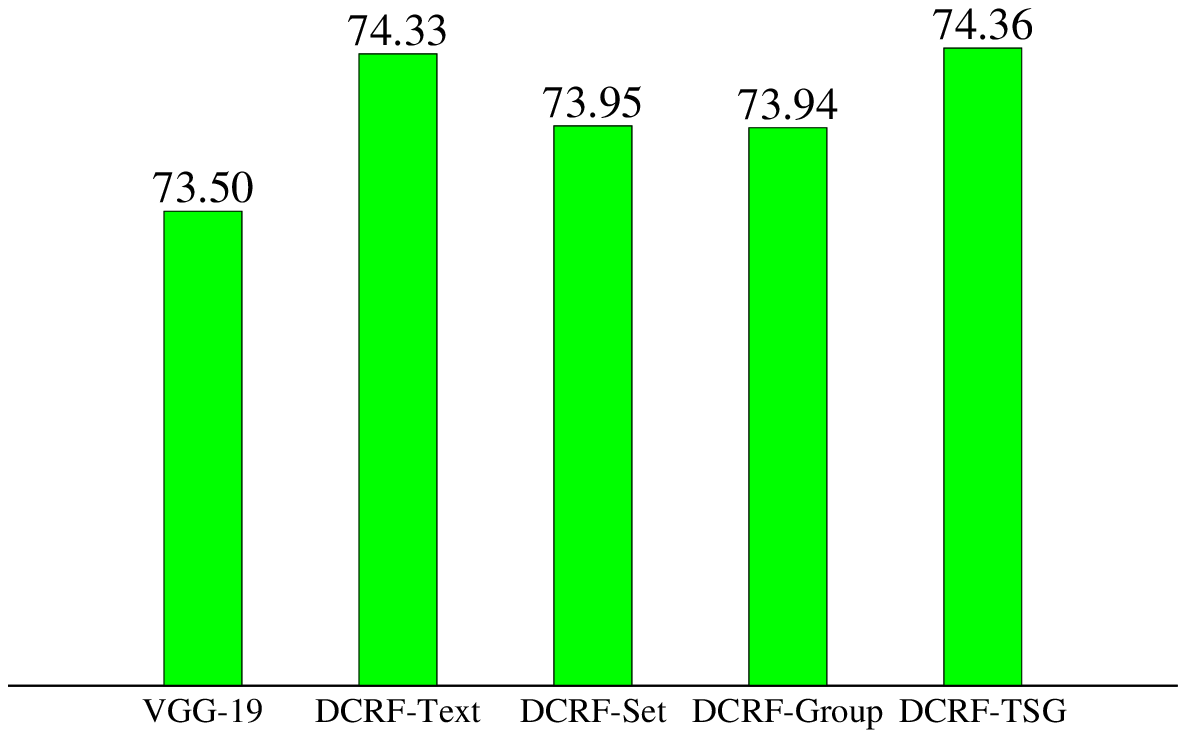}}
\end{center}
\end{minipage}
\begin{minipage}{0.230\textwidth}
\begin{center}
\subfloat[AP with ResNet-152]{\label{fig:metacnnres_ap}
\includegraphics[height=0.58\textwidth, width=0.20\textheight, angle=0]{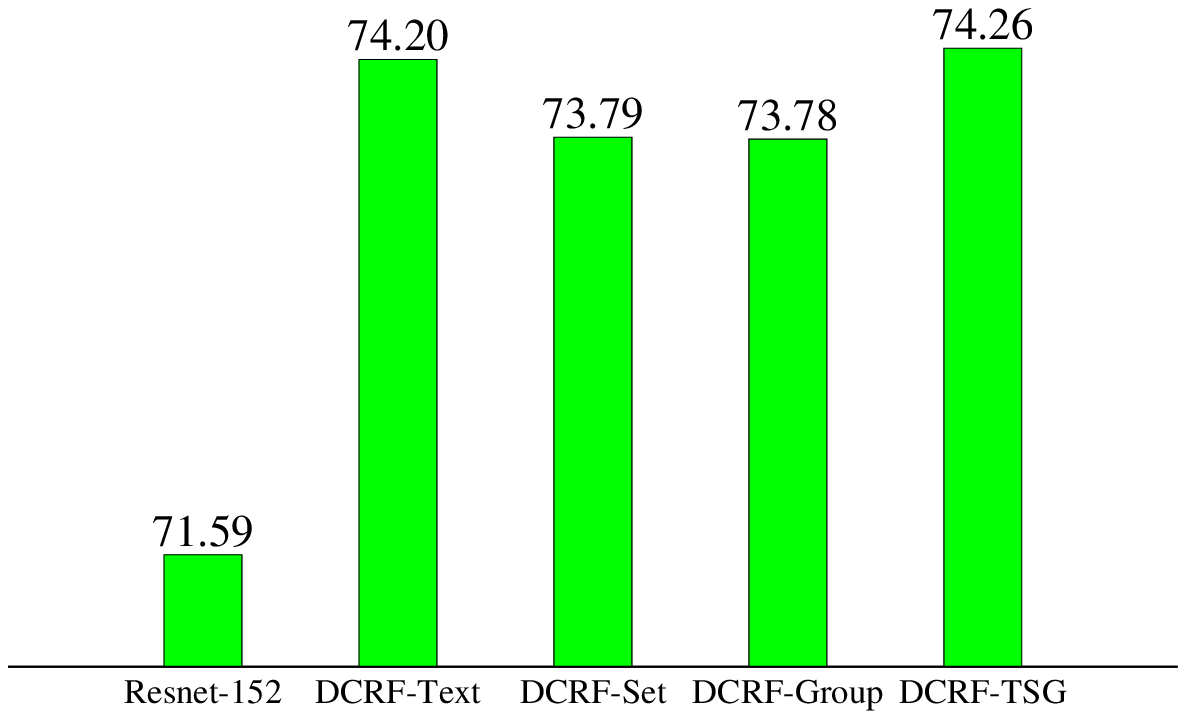}}
\end{center}
\end{minipage}
\begin{minipage}{0.230\textwidth}
\begin{center}
\subfloat[Recall with VGG-19]{\label{fig:metacnnvgg_rec}
\includegraphics[height=0.58\textwidth, width=0.20\textheight, angle=0]{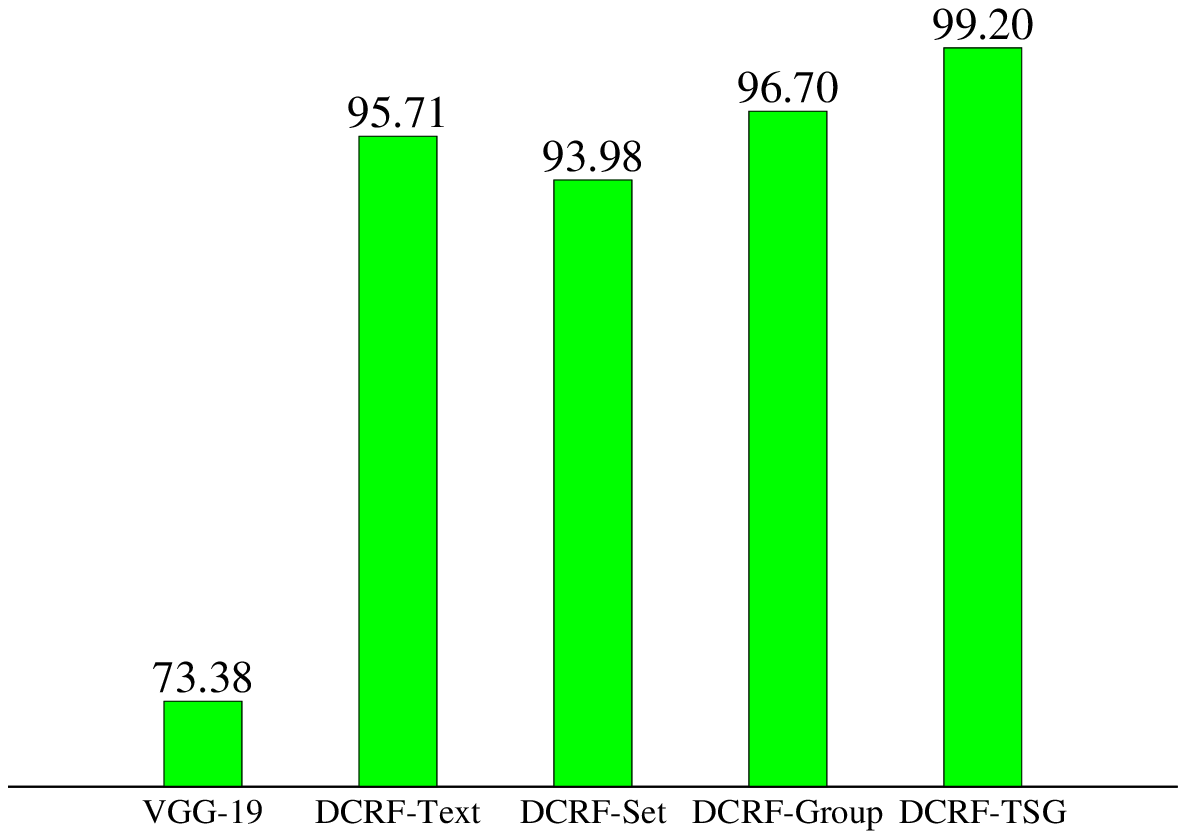}}
\end{center}
\end{minipage}
\begin{minipage}{0.230\textwidth}
\begin{center}
\subfloat[Recall with ResNet-152]{\label{fig:metacnnres_rec}
\includegraphics[height=0.58\textwidth, width=0.20\textheight, angle=0]{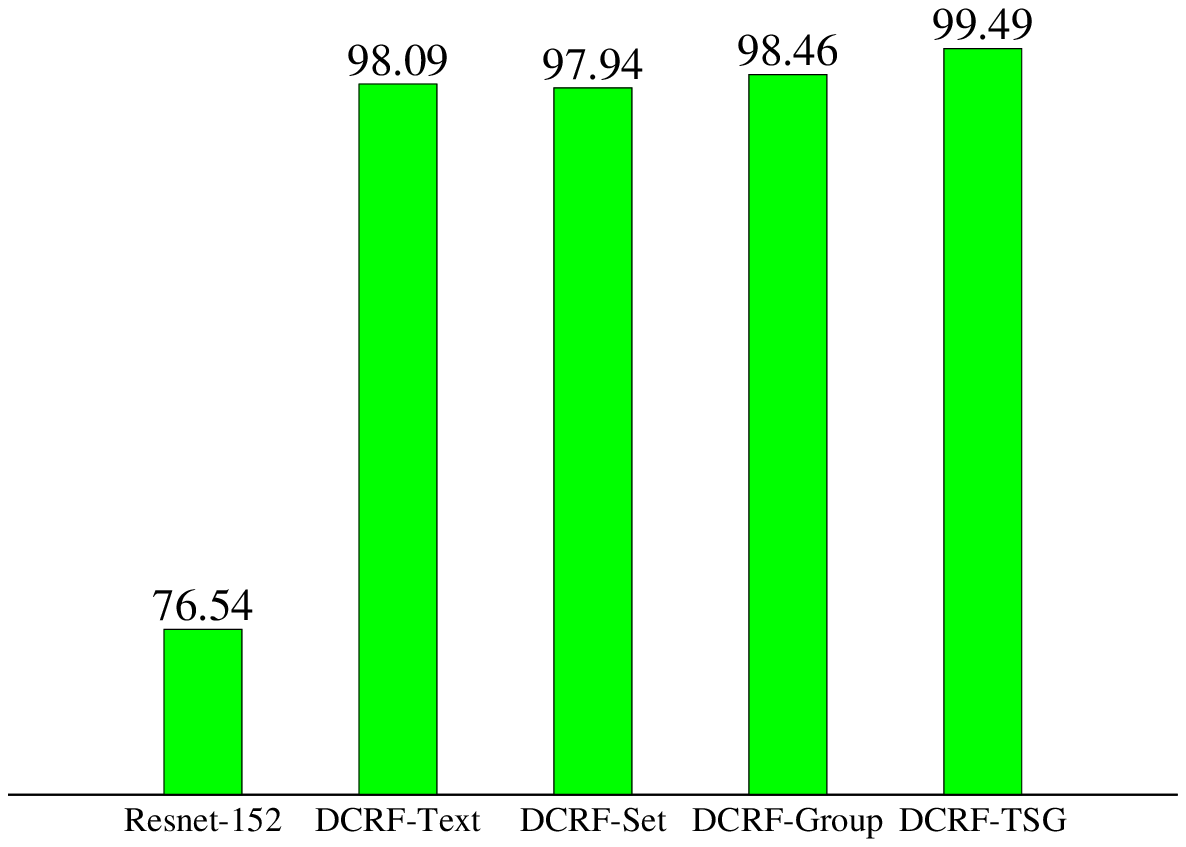}}
\end{center}
\end{minipage}
\begin{minipage}{0.230\textwidth}
\begin{center}
\subfloat[Precision with VGG-19]{\label{fig:tagcnnvgg_pre}
\includegraphics[height=0.58\textwidth, width=0.20\textheight, angle=0]{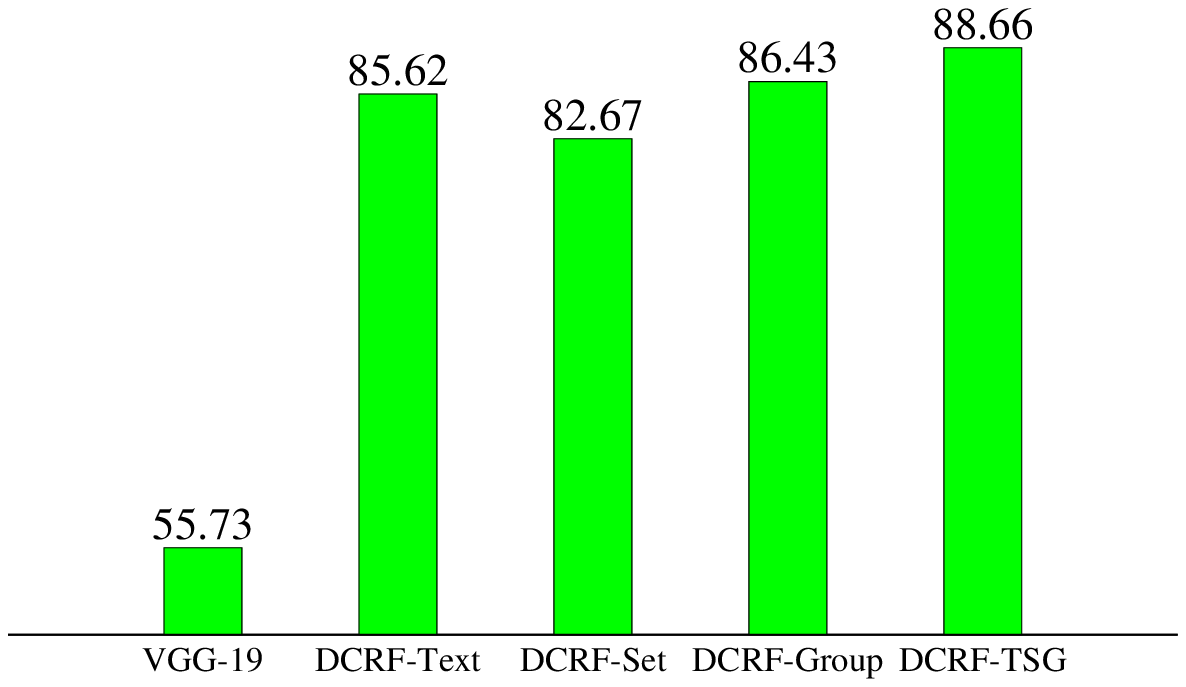}}
\end{center}
\end{minipage}
\begin{minipage}{0.230\textwidth}
\begin{center}
\subfloat[Precision with ResNet-152]{\label{fig:metacnnres_pre}
\includegraphics[height=0.58\textwidth, width=0.20\textheight, angle=0]{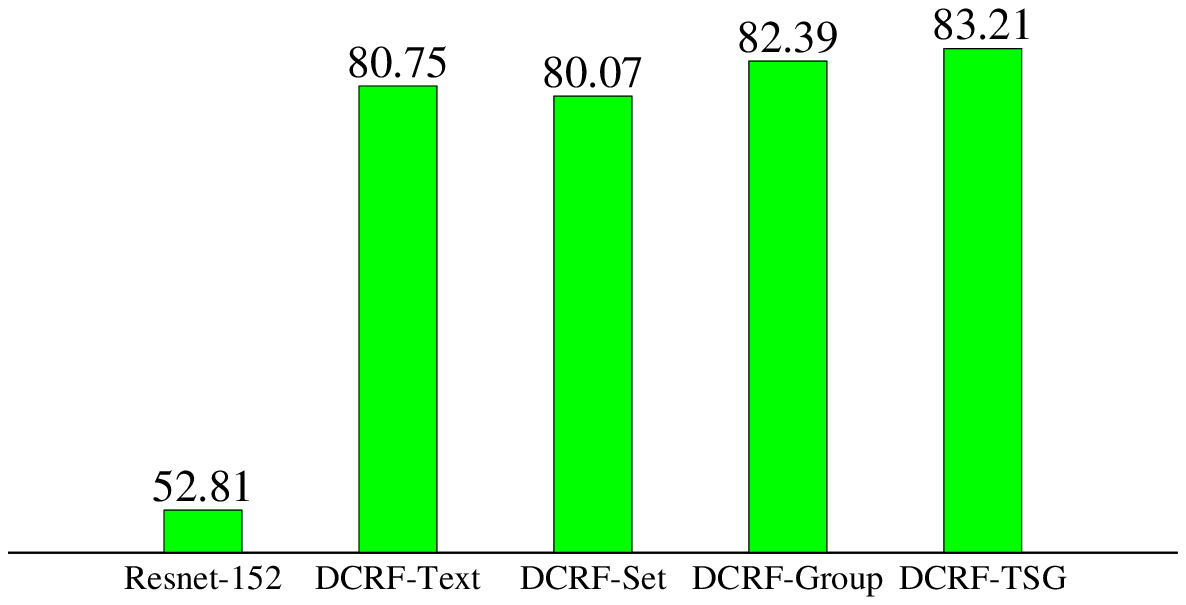}}
\end{center}
\end{minipage}
\begin{minipage}{0.240\textwidth}
\begin{center}
\subfloat[Accuracy with VGG-19]{\label{fig:metacnnvgg_acc}
\includegraphics[height=0.58\textwidth, width=0.20\textheight, angle=0]{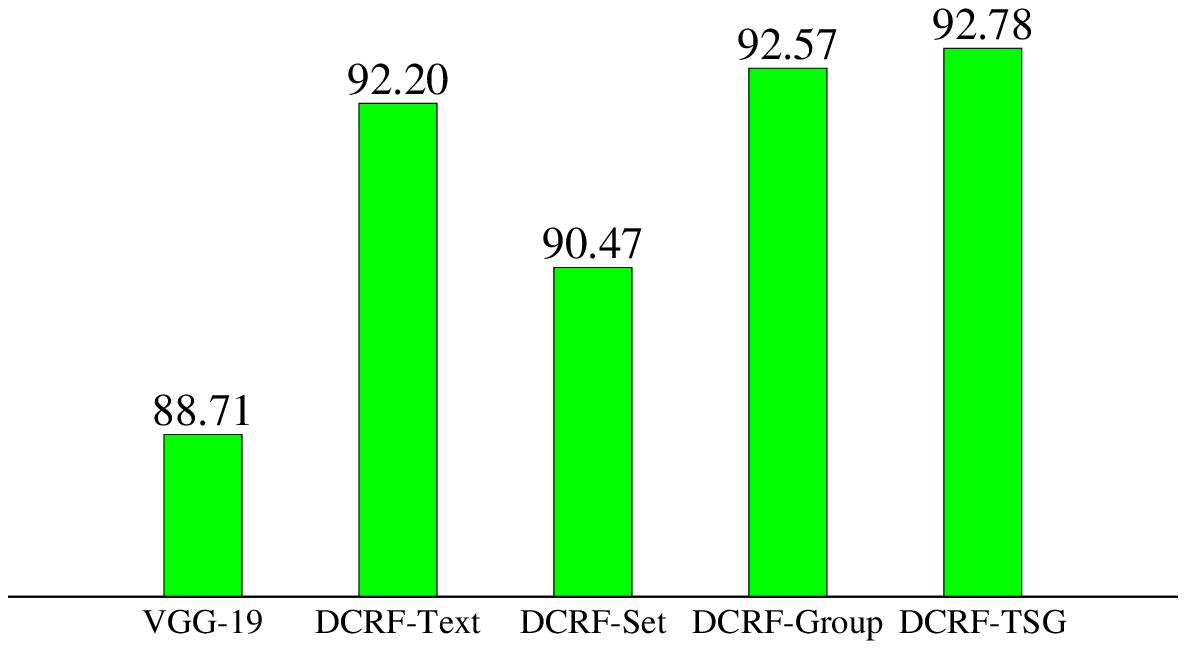}}
\end{center}
\end{minipage}
\begin{minipage}{0.230\textwidth}
\begin{center}
\subfloat[Accuracy with ResNet-152]{\label{fig:metacnnres_acc}
\includegraphics[height=0.58\textwidth, width=0.20\textheight, angle=0]{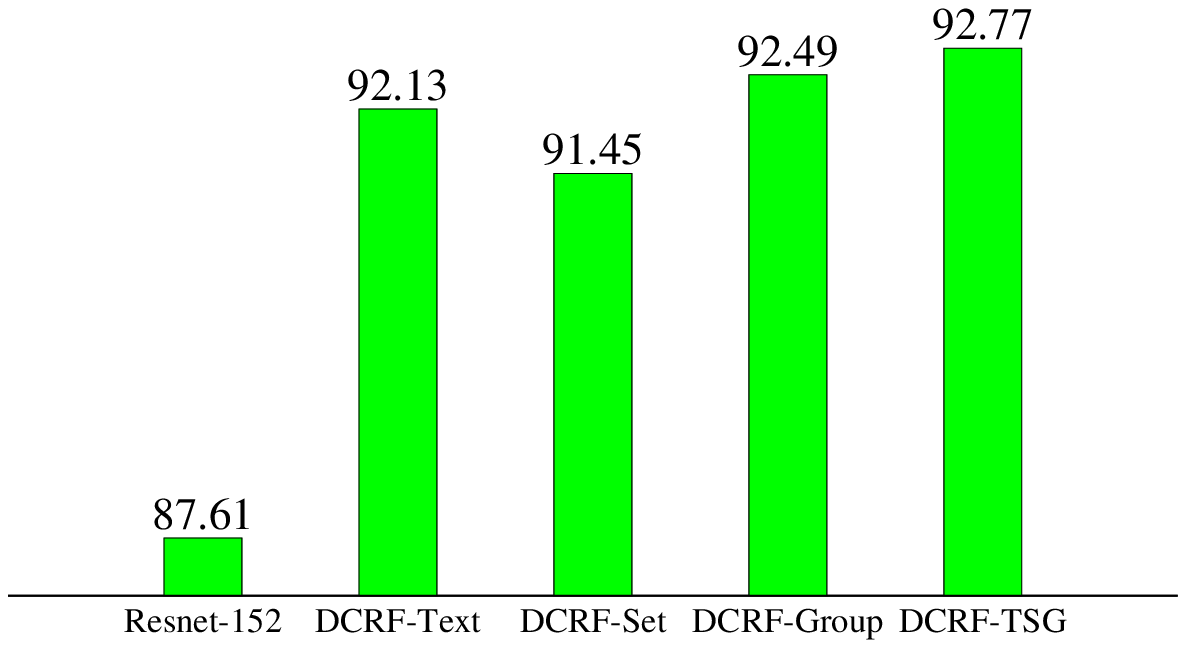}}
\end{center}
\end{minipage}
\vspace{-0.10in}
\caption{The comparison result with node features extracted from the VGG-19 and ResNet-152 networks (unit: \%).}\label{fig:meta_cnn}
\end{figure}
 
We summarize the results in Figure~\ref{fig:meta_cnn}. As we expect, each of all these DCRF-Text, DCRF-Set and DCRF-Group perform better than VGG-19 and ResNet-152, which suggests that all these three types of metadata are helpful for image annotation. Among these three types of metadata, using text information provides the greatest improvement in AP compared to the other two. However, 
combining all three types into the DCRF-TSG model produces the greatest performance in all four metrics, regardless of using the VGG-19 or ResNet-152 network as node feature extractor. Such observations demonstrate that the metadata as text, user sets and image groups are complementary to each other and can be used for boosting the quality of image labeling.


\subsection{Compare with state-of-the-art approach}
In addition to the text-level CNN model $\text{CNN}_{text}$~\cite{Kim14f}, and four image-level popular CNN models $\text{AlexNet}_{img}$~\cite{AlexNIPS2012}, $\text{VGG-19}_{img}$~\cite{Simonyan14c}, $\text{ResNet-152}_{img}$~\cite{He_2016_CVPR} and $\text{DenseNet-201}_{img}$~\cite{huang2017densely}, we compare our proposed DCRF with the most closely related work, {\em i.e.}, McAuley {\em et al}'s CRF algorithm~\cite{McAuleyECCV12}, denoted as McAuley-CRF, which explores the social-network metadata such as image groups and comments, and utlizes structured learning techniques to learn model parameters. We also compare with one deep learning related work, {\em i.e.}, Johnson {\em et al}'s neighbor-based CNN algorithm~\cite{Johnson:ICCV15}, denoted as Johnson-NCNN, which use image metadata in a nonparametrical manner to generate neighborhoods of related images using Jaccard similarity and then uses a deep learning to blend visual information from the image and its neighbors. 
\begin{table}[h]
   \vspace{-0.07in}
   \caption{The performance comparison among the competing algorithms (AP: average precision, REC: recall, PRE: precision, ACC: accuarcy, unit: \%).}
   \vspace{-0.10in}
   \begin{center}
      \setlength{\tabcolsep}{4.9pt}
      \begin{tabular}{c|c c c c}
       \hline	
	 & AP  & REC  &	PRE  &	ACC \\
       \hline
       $\text{CNN}_{text}$~\cite{Kim14f} &	27.97 & 25.39 & 32.76 & 82.47 \\
       \hline
       $\text{AlexNet}_{img}$~\cite{AlexNIPS2012} &	62.54	& 76.30 & 40.25 & 74.56 \\
       $\text{VGG-19}_{img}$~\cite{Simonyan14c} &	{\bf 73.50}	& {\bf 77.38} & {\bf 55.73} & {\bf 88.71} \\
       $\text{ResNet-152}_{img}$~\cite{He_2016_CVPR} &	71.59	& 76.54 & 52.82 & 87.62 \\
       $\text{DenseNet-201}_{img}$~\cite{huang2017densely} &	63.26	& 72.55 & 42.93 & 85.06 \\
       \hline
       McAuley-CRF~\cite{McAuleyECCV12} &	54.73	& 40.75 & 59.44 & 83.1 \\
       $\text{John-NCNN}_{vgg}$~\cite{Johnson:ICCV15} &  {\bf 73.78} & {\bf 61.18} & 79.01 & {\bf 92.57} \\	
       $\text{John-NCNN}_{res}$~\cite{Johnson:ICCV15} &	72.90	& 50.59 & {\bf 81.39} & 91.87 \\
      \hline
      $\text{DCRF}_{vgg}$-BCE & 74.13 & 92.66 & 85.86 & 92.50\\
      $\text{DCRF}_{vgg}$-RLoss & 74.29 & 93.12 & 88.18 & 92.61\\  
      $\text{DCRF}_{vgg}$-BCE+RLoss & {\bf 74.36} & 99.20 & {\bf 88.66} & {\bf 92.78}\\
      $\text{DCRF}_{res}$-BCE & 74.05 & 91.52 & 74.69 & 91.74\\
      $\text{DCRF}_{res}$-RLoss & 74.09 & 94.38 & 77.59 & 91.93\\
      $\text{DCRF}_{res}$-BCE+RLoss & 74.26 & {\bf 99.49} & 83.21 & 92.77\\
      \hline
   \end{tabular}
   \vspace{-0.35cm}
\end{center}
\label{tab:comparision}
\end{table}

For fair comparison, we provide two versions of deep learning models ({\em i.e.}, the VGG-19 and ResNet-152 networks) for Johnson-NCNN, and mark them as $\text{Johnson-NCNN}_{vgg}$ and $\text{Johnson-NCNN}_{res}$, respectively. Obviously, our DCRF has two versions, {\em i.e.}, $\text{DCRF}_{vgg}$ and $\text{DCRF}_{res}$. To better show the effectiveness of the loss function we use in Section~\ref{sec:learnandinference}, we get six versions, {\em i.e.}, $\text{DCRF}_{vgg}$-BCE,  $\text{DCRF}_{res}$-BCE, $\text{DCRF}_{vgg}$-RLoss,  $\text{DCRF}_{res}$-RLoss, $\text{DCRF}_{vgg}$-BCE+RLoss,  $\text{DCRF}_{res}$-BCE+RLoss, in which ``BCE" indicates binary cross entropy and ``RLoss" means pairwise ranking loss in the binary classification cases.

\begin{figure*}
\captionsetup[subfigure]{labelformat=empty}
\begin{minipage}{0.245\textwidth}
\begin{center}
\subfloat[]{\label{fig:visexample1}
\includegraphics[height=0.88\textwidth, width=0.19\textheight, angle=0]{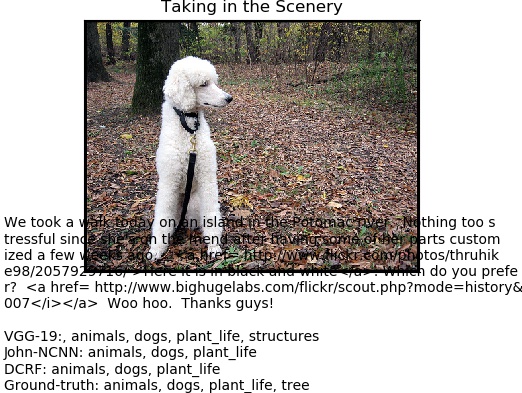}}
\end{center}
\end{minipage}
\begin{minipage}{0.245\textwidth}
\begin{center}
\subfloat[]{\label{fig:visexample2}
\includegraphics[height=0.88\textwidth, width=0.19\textheight, angle=0]{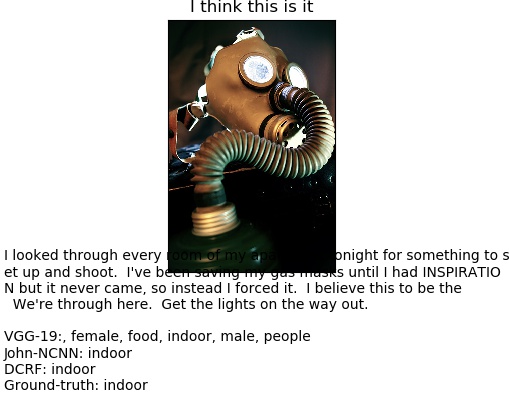}}
\end{center}
\end{minipage}
\begin{minipage}{0.245\textwidth}
\begin{center}
\subfloat[]{\label{fig:visexample3}
\includegraphics[height=0.88\textwidth, width=0.19\textheight, angle=0]{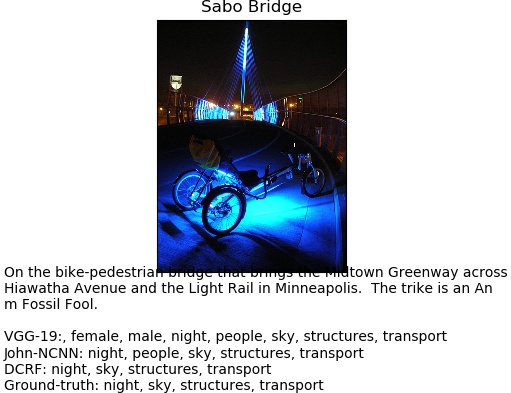}}
\end{center}
\end{minipage}
\begin{minipage}{0.245\textwidth}
\begin{center}
\subfloat[]{\label{fig:visexample4}
\includegraphics[height=0.88\textwidth, width=0.19\textheight, angle=0]{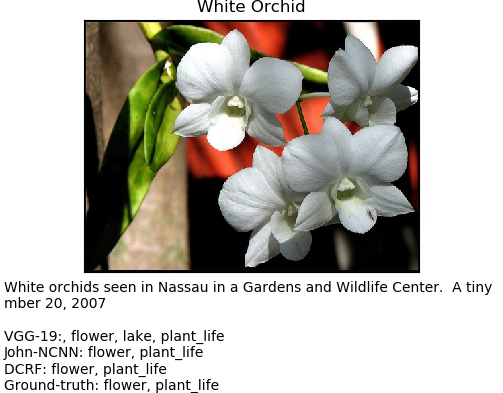}}
\end{center}
\end{minipage}\\
\begin{minipage}{0.245\textwidth}
\begin{center}
\subfloat[]{\label{fig:visexample5}
\includegraphics[height=0.88\textwidth, width=0.19\textheight, angle=0]{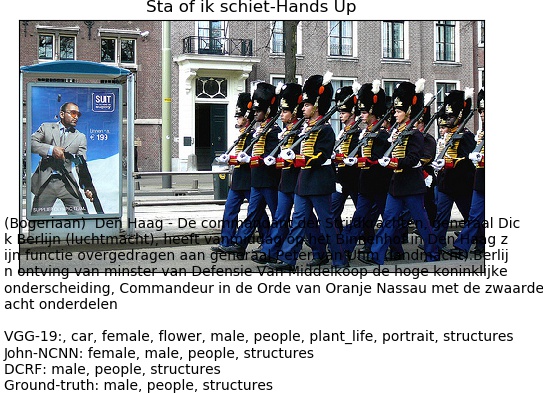}}
\end{center}
\end{minipage}
\begin{minipage}{0.245\textwidth}
\begin{center}
\subfloat[]{\label{fig:visexample6}
\includegraphics[height=0.88\textwidth, width=0.19\textheight, angle=0]{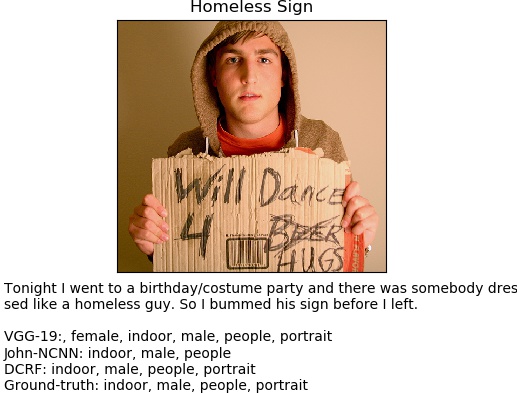}}
\end{center}
\end{minipage}
\begin{minipage}{0.245\textwidth}
\begin{center}
\subfloat[]{\label{fig:visexample7}
\includegraphics[height=0.88\textwidth, width=0.19\textheight, angle=0]{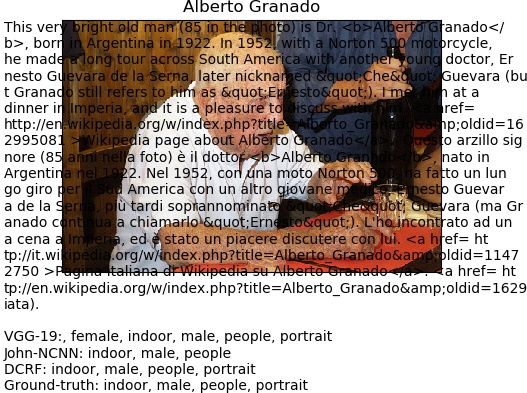}}
\end{center}
\end{minipage}
\begin{minipage}{0.245\textwidth}
\begin{center}
\subfloat[]{\label{fig:visexample8}
\includegraphics[height=0.88\textwidth, width=0.19\textheight, angle=0]{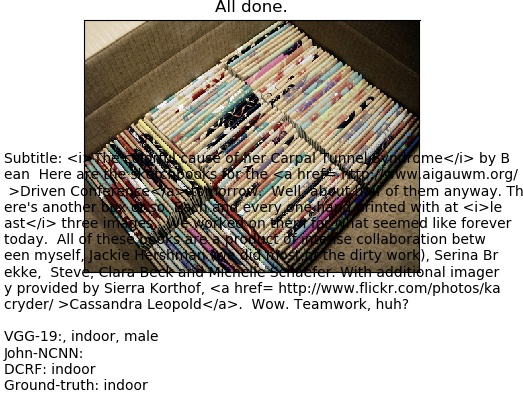}}
\end{center}
\end{minipage}\\
\begin{minipage}{0.245\textwidth}
\begin{center}
\subfloat[]{\label{fig:visexample9}
\includegraphics[height=0.88\textwidth, width=0.19\textheight, angle=0]{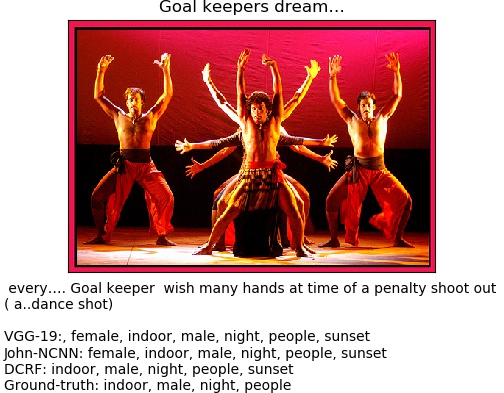}}
\end{center}
\end{minipage}
\begin{minipage}{0.245\textwidth}
\begin{center}
\subfloat[]{\label{fig:visexample10}
\includegraphics[height=0.88\textwidth, width=0.19\textheight, angle=0]{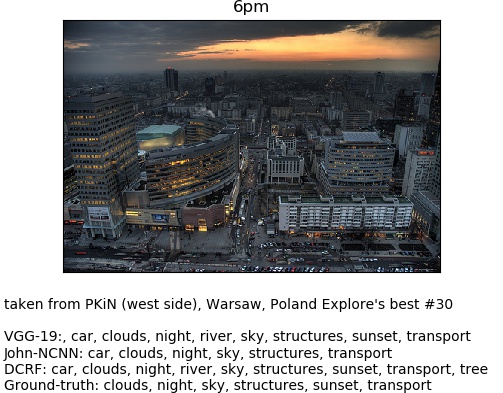}}
\end{center}
\end{minipage}
\begin{minipage}{0.245\textwidth}
\begin{center}
\subfloat[]{\label{fig:visexample11}
\includegraphics[height=0.88\textwidth, width=0.19\textheight, angle=0]{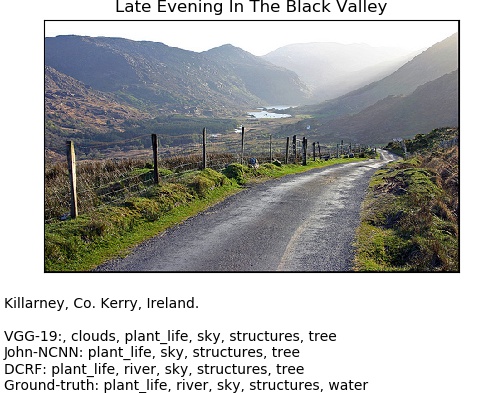}}
\end{center}
\end{minipage}
\begin{minipage}{0.245\textwidth}
\begin{center}
\subfloat[]{\label{fig:visexample12}
\includegraphics[height=0.88\textwidth, width=0.19\textheight, angle=0]{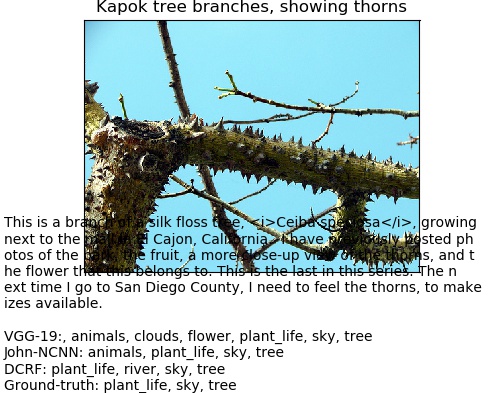}}
\end{center}
\end{minipage}\\
\vspace{-0.20in}
\caption{Visualization of image labeling on some testing examples. For each example, the tile is above the image, and the text below the image is the corresponding description. The bottom four rows are prediction labels by VGG-19, $\text{John-NCNN}_{vgg}$, $\text{DCFR}_{vgg}$ and the corresponding groud-truth labels.}\label{fig:visualization}
\vspace{-0.20in}
\end{figure*}

The results are summarized in Table~\ref{tab:comparision}, from which we can observe: (a) all four image-level CNNs perform better $\text{CNN}_{text}$, and VGG-19 and ResNet-152 are the top 2 image-level CNN models on the MIR-9K dataset; (b) our proposed DCRF significantly outperforms the McAuley-CRF approach, which shows the big advantages of using deep neural networks in CRF; (c) with the same deep node features extracted from either the VGG-19 or ResNet-152 network, all versions of our DCRF are able to obtain improvement in all four metrics when compared to using textual or visual information only; 
(d) rank loss function works a little better than cross entropy loss function in all four metrics; 
(e) the performance of our $\text{DCRF}_{vgg}$-BCE+RLoss and $\text{DCRF}_{res}$-BCE+RLoss are slightly higher than those of John-$\text{NCNN}_{vgg}$ and John-$\text{NCNN}_{res}$ in AP and accuracy and significantly higher in recall and precision, and our $\text{DCRF}_{vgg}$-BCE+RLoss achieves the best performances in AP, precision and accuracy. 

\subsection{Visualization}
For better understanding of our porposed DCRF, we visualize some testing examples with $\text{DCRF}_{vgg}$ and take VGG-19 and John-$\text{NCNN}_{vgg}$ as baselines in Figure~\ref{fig:visualization}.
As we can see, 
both $\text{DCRF}_{vgg}$ and John-$\text{NCNN}_{vgg}$ benefit from the metadata information for improving the quality of image labeling. Overall, our proposed DCRF achieves the higher quality of labels. Moreover, our proposed $\text{DCRF}_{vgg}$ is able to predict some categories that are not clear or even occluded in images, such as ``car" and ``tree" in the middle two examples at the bottom row.

\vspace{-0.05cm}
\section{Conclusion}
In this paper, we propose a novel deep fully connected CRF based framework DCRF with deep neural networks for image labeling using social network metadata. In such a framework, CNNs are used to extract powerful visual features for nodes/images and textual features to explore the underlying information embedded in text. The fully connected CRF graph is built based on the textual similarity and the link-based metadata like user sets and image groups. With the mean-field approximation modeled as an RNN, our proposed framework DCRF becomes a joint end-to-end CNN-RNN formulation, which combines the strengths of both CNNs and RNNs. The experimental evaluation on the MIR-9K dataset demonstrates that our proposed DCRF framework outperforms state-of-the-art approaches~\cite{McAuleyECCV12, Johnson:ICCV15}.

Our future work includes investigating more effective meta information, and improving the efficiency of the current DCRF framework to handle more complicated real-world application problems.

\newpage
{\small
\bibliographystyle{ieee}
\bibliography{DNNCRF_Arxiv}
}

\end{document}